\documentclass{article}

\usepackage[preprint]{neurips_2025}

\usepackage[utf8]{inputenc} 
\usepackage[T1]{fontenc}    
\usepackage{hyperref}       
\usepackage{url}            
\usepackage{booktabs}       
\usepackage{amsfonts}       
\usepackage{nicefrac}       
\usepackage{microtype}      
\usepackage{xcolor}         
\usepackage{amsmath}
\usepackage{graphicx}
\usepackage{subcaption}
\usepackage{algorithm}
\usepackage{algpseudocode}
\usepackage{comment}
\usepackage{wrapfig}
\usepackage{tikz}
\usepackage{enumitem}
\usepackage{amssymb}
\usepackage{pifont}
\usepackage{tablefootnote}
\newcommand{\xmark}{\ding{55}}
\usepackage{array}
\newcolumntype{C}[1]{>{\centering\arraybackslash}p{#1}}
\title{Scaling Context Requires Rethinking Attention}

\author{
  Carles Gelada\footnotemark[1] \\
  Manifest AI \\
  \And
  Jacob Buckman\footnotemark[1] \\
  Manifest AI \\
  \And
  Sean Zhang\footnotemark[1] \\
  Manifest AI \\
  \And
  Txus Bach \\
  Manifest AI
}

\begin{document}

\begingroup
\microtypesetup{expansion=false}
\renewcommand{\thefootnote}{\fnsymbol{footnote}}

\footnotetext[1]{Equal contribution. Correspondence to: \texttt{cgel.saez@gmail.com}, \texttt{jacobbuckman@gmail.com}, \texttt{seanxwzhang@gmail.com} }

\maketitle
\endgroup               
\setcounter{footnote}{0}  

\newcommand{\tpow}{\textsc{tpow}}
\newcommand{\spow}{\textsc{spow}}
\newcommand{\tiledspow}{\textsc{tspow}}
\newcommand{\R}{\mathbb{R}}
\newcommand{\N}{\mathbb{N}}
\newcommand{\mi}{\N_d^{\times p}}
\newcommand{\iMI}{(i_1,\cdots,i_p)}
\newcommand{\dtile}{d{\text-tile}}
\newcommand{\ndmi}{{\text{NDMI}^p_d}}
\newcommand{\tiledndmi}{{\text{NDMI}^p_{d/\dtile}}}
\newcommand{\hist}{\text{hist}}
\newcommand{\attn}{\text{attn}}
\newcommand{\expattn}{\attn_\text{exp}}
\newcommand{\linattn}{\attn_\text{lin}^\phi}
\newcommand{\linattntpow}{\attn_\text{lin}^{\tpow_p}}
\newcommand{\powattn}{\attn_{\text{pow}}^p}
\newcommand{\cmark}{\ding{51}}            
\newcommand{\bcmark}{{\bfseries\ding{51}}} 

\newtheorem{theorem}{Theorem}[section]
\newtheorem{lemma}[theorem]{Lemma}

\begin{abstract}

We argue that neither transformers nor sub-quadratic architectures are well suited to training at long sequence lengths: the cost of processing the context is too expensive in the former, too inexpensive in the latter.
Approaches such as sliding window attention which reduce the cost-per-token of a transformer impair in-context learning, and so are also unsuitable.
To address these limitations, we introduce \textit{power attention}, an architectural layer for linear-cost sequence modeling whose state size can be adjusted independently of parameters, unlocking the advantages of linear attention on practical domains.
We develop and open-source a set of GPU kernels for efficient power attention, identifying a novel pattern of operation fusion to avoid memory and bandwidth bottlenecks.
Our experiments on the in-context learning of power attention shows that these models dominate both exponential attention and linear attention at long-context training.

\end{abstract}

\section{Introduction}

Many techniques to improve the performance of language models involve adding tokens to the context. One popular approach is to include reference material, such as by adding the content of a codebase to the context of a coding assistant \citep{swebench}. Another approach is to introduce tokens sampled from the model itself, as is done by chain-of-thought LLMs \citep{deepseek-r1, wei2022chain}. A third approach is to use LLM agents, which iteratively interact with the world via tool use and adapt to feedback via context tokens \citep{yang2024swe, he2024webvoyager, schick2023toolformer}. If these context scaling techniques continue to pay off, one might expect a future where contexts regularly contain millions or even billions of tokens.

However, it remains unclear what architectures are best suited for training with long contexts. It is commonly argued that, despite their ubiquity, transformers \citep{attention_is_all_you_need} are poorly suited to long-context training due to their use of self-attention, whose compute cost grows quadratically with context length. The fact that modern transformer-based LLMs are trained primarily on context lengths between 4k and 32k tokens \citep{llama3,llama4,gemini}, with long-context training relegated to post-training (if at all), lends credence to this position. These concerns have motivated research on so-called \textit{subquadratic sequence architectures} such as those proposed by \citet{retnet, rwkv, mamba}. These architectures primarily utilize variants of \textit{linear attention}, an operation similar to the attention layer of transformers except that it allows for a recurrent linear-cost formulation.

In Section \ref{sec:balance} we argue that any strong long-context architecture must possess three attributes:
\begin{enumerate}[noitemsep, topsep=0pt]
    \item A balanced weight-to-state ratio at long contexts.
    \item Admits an efficient hardware-aware implementation on tensor cores.
    \item Good in-context learning (ICL) ability.
\end{enumerate}
We then show that neither attention-based architectures, nor existing subquadratic architectures, meet these criteria. Table~\ref{tab:arch_comparison} summarizes our perspective.

In Section \ref{sec:powatt} we introduce \textit{power attention}, a powerful variant of linear attention. Power attention possesses a hyperparameter $p$ which controls the state size independently of the parameter count, enabling us to balance the weight-state FLOPs ratio for architectures of any scale. It also admits a hardware-aware implementation for training on GPUs. Section \ref{sec:kernels} describes the implementation of our open-source kernels, which enable real wall-clock speedups over Flash Attention in practical settings (e.g. $p=2$ is 8.6x faster at 64k context). Furthermore, our kernels still lag behind Flash Attention \citep{flash2} in terms of hardware utilization, and so we expect future engineering efforts which close this gap to result in even larger speedups.

\begin{wraptable}{r}{0.55\textwidth}
\centering
\begin{tabular}{|c|c|c|c|}
\hline
\textsc{Architecture} & \small\textsc{Balance} & \small\textsc{Efficiency} & \small\textsc{ICL} \\
\hline
Transformer & \xmark\quad & \checkmark & \checkmark \\
Classic RNNs & \xmark\quad & \xmark\quad & \checkmark \\
Modern RNNs & \xmark\quad & \checkmark & \checkmark \\
Windowed Attention & \checkmark & \checkmark & \xmark \\
\textbf{Power Attention} & \checkmark & \checkmark & \checkmark \\
\hline
\end{tabular}
\vspace{0.5em}
\caption{
Comparison of approaches. Section~\ref{sec:balance} justifies the importance of these criteria, and explains why each architecture passes or fails.
}
\label{tab:arch_comparison}
\end{wraptable}

We evaluate power attention empirically in Section~\ref{sec:experiments}. Experiments in Section~\ref{sec:experiments_icl} show that power attention has better in-context learning than other balanced architectures.
In Section~\ref{sec:long_context_training}, we show that when training on contexts of length 65536, power attention dominates both exponential and linear attention in terms of loss-per-FLOP.

These results have two main limitations.
Firstly, our experiments are limited to measuring negative log likelihood on a dataset of generic natural language text. We did not study other domains, modalities, or downstream tasks. Secondly, in our setting, the compute-optimal context grows relatively slowly, diminishing the value of long-context training. We leave to future work the replication of these results across a variety of settings and metrics, and exploration to identify domains with long compute-optimal contexts (perhaps tasks that require chain-of-thought reasoning, or modalities such as audio).

\section{Background}
\label{sec:background}

\textbf{Sequence modeling.} Let $\mathcal{X}$ denote a finite set of tokens, referred to as the \textit{vocabulary}. Let $\mathcal{X}^t$ denote the set of length-$t$ sequences over $\mathcal{X}$, the \textit{documents}.  Given some distribution $\mathbb{D} \in \text{Dist}(\mathcal{X}^t)$ we are concerned with finding a model assigning the maximum probability to documents sampled from $\mathbb{D}$. A common approach is \textit{causal sequence modeling}, based on a model $f_\theta$ mapping sequences $\mathcal{X}^i$ of arbitrary length $i \in \mathbb{N}$ to distributions over next-tokens, $\text{Dist}(\mathcal{X})$, where $\theta \in \Theta$ denotes the \textit{parameters} of the model. Implicitly, such a model defines a distribution over the space of all documents $x \in \mathcal{X}^t$ via the autoregressive factorization:
\begin{align}
f_\theta(x) = f_\theta(x_1, \dots, x_t) = \prod_{i=1}^t f_\theta(x_i \mid x_{<i})
\end{align}
The goal of causal sequence modeling is to learn parameters $\theta$ such that the induced distribution $f_\theta$ matches the data distribution, where error is typically measured by the \textit{cross-entropy loss}:
\begin{align}
\mathcal{L}_D(\theta) = \mathbb{E}_{x \sim \mathbb{D}} \left[ -\log f_\theta(x) \right]
\end{align}

\paragraph{Recurrent neural networks.} RNNs \citep{classic_rnns, lstm, gru} are models $f_\theta$ which can be expressed using a Markovian \textit{state}, $S_i\in \R^n$, which summarizes the information of the entire input history $x_{\le i} = x_1, \cdots, x_{i}$. The output of an RNN can be expressed as $y_{i} = g_\theta(x_i, S_i)$ and the state evolves according to a recurrent relation $S_{i+1} = h_\theta(x_i, S_i)$.

\paragraph{Attention.} The causal self-attention layer, a critical piece of the \textit{transformer} architecture \citep{attention_is_all_you_need}, is defined as follows. Let $Q,K \in \R^{t\times d}, \; V \in \R^{t\times v}$ be the query, key and value matrices. We can also think of them as sequences of vectors $Q_i, K_i \in \R^d$ and $V_i\in \R^v$. The output of the attention layer is a matrix $ \expattn(Q,K,V) \in \R^{t\times v}$ defined as
\begin{align}
\expattn(Q,K,V)_i = \sum_{j=1}^i e^{Q^T_i K_j} V_j
\end{align}
This can be implemented efficiently in matrix form by using a mask $M \in \R^{t \times t}$ where $M_{ij} = \mathbf{1}_{i \leq j}$,
\begin{align}
\expattn(Q,K,V) = \left( \text{exp} ( Q K^T )  \odot M \right)  V
\end{align}
where $\exp(A)$ denotes element-wise exponentiation of the matrix $A$.

Attention can be expressed in an RNN-like form. The outputs $Y_i$ depend only on a state $S_i = \left ( K_{\leq i}, V_{\leq i} \right ) \in \R^{t\times d} \oplus  \R^{t\times v}$, commonly called the \textit{KV cache}. The main difference from conventional RNNs is that the state of attention does not have a fixed dimensionality; it grows with sequence length.

\paragraph{Normalization.} To stabilize learning, attention usually requires normalization. The original (and most common) approach to normalization is to divide by the sum of the attention scores, turning them into a probability distribution \citep{attention_is_all_you_need}. We use this normalization throughout this work. One limitation of this approach is that it requires positive attention scores. Other approaches have been proposed \citep{attention_sink,sigmoid_attention}, but we do not consider them.

\textbf{Sliding window attention.} A variant of attention which chooses a window size $w$, and truncates the KV cache to this length using a first-in-first-out approach \citep{sparse_transformers}. The formula for the outputs is $\sum_{j=i-w}^i e^{Q^T_i K_j} V_j$.

\textbf{Linear attention.} \citet{katharopoulos-et-al-2020} removes the exponential from attention and projects the keys and queries using $\phi: \R^d \to \R^D$,
\begin{align}
\linattn(Q,K,V) = \left ( \phi(Q) \phi(K)^T \odot M \right) V
\end{align}
where $\phi(A) \in \R^{t\times D}$ denotes application of $\phi$ to the rows of $A \in \R^{t\times d}$.
The key property of linear attention is that it admits an alternative to the KV cache, a constant-size state $S_i \in \R^{v\times D}$ unrolled via the recurrence relation:
\begin{align}
\linattn(Q,K,V)_i =  S_i \phi(Q_i)   \qquad
 S_i =  S_{i-1} + V_i  \phi(K_i)^T
\end{align}
The array of $t$ outputs can be computed with cost $O(tDv)$. For this reason, for long sequences, the \textit{recurrent form} is preferred over the \textit{attention form} on the KV cache, which has cost $O\left(t^2(d+v)\right )$. This recurrent form also highlights the motivation behind the inclusion of $\phi$. Since $S \in \mathbb{R}^{Dv}$, the choice of $\phi$ can be used to adjust the state size, known as \textit{state expansion} \citep{linear_transformer_fast_weight_programmers}.

\paragraph{Chunked form.}  The recurrent form of linear transformers is rarely useful in practice. The states $S_i \in \R^{v\times D}$ are typically large, so having to compute and store in memory every state in the sequence becomes a major bottleneck. The \textit{chunked form} \citep{buckman2024,retnet} interpolates between the recurrent form and the attention form, capturing benefits of both. The key idea is to compute only a subset of all states: $S_0, S_c, S_{2c}, \cdots$, for some appropriately chosen chunk size $c \in \N$. The chunked form is given by the following equation:
$$
Y_{nc+m} = S_{nc} \phi(Q_{nc+m}) + \sum_{j=nc+1}^{nc+m} (Q_{nc+m}^T K_j) V_j 
$$
For any $i$ there exist $0 \le n$ and $0\le m < c$ such that $i=nc+m$. So $Y_{nc+m}$ can be computed with an interaction with the state $S_{cn}$ of cost $O(vD)$ and an intra-chunk attention of cost $O(cd)$. Thus, the cost of the entire output sequence is $O(tDv + tcd)$. 

\paragraph{Gating.} On long-context tasks, it is common to give a mechanism for the network to directly avoid attending to old data. Originally, this was done at a fixed rate using techniques such as ALiBi \citep{alibi}. More recently, \citet{forgetting_transformer} propose a learned gating value per timestep, which is the approach we adopt in this work. Gating has been demonstrated to be particularly important in linear attention \citep{gsa,gla,mamba}.

\paragraph{Architectures.} Self-attention layers are merely one piece of a broader transformer architecture, which typically alternates between attention and MLP layers \citep{attention_is_all_you_need}. Modern architectures often also include components such as rotary embeddings \citep{rope} and local convolutions \citep{gdn}. In this work, we focus our study only on the attention layer, and in general do not modify other architectural components. We use the FLA codebase \citep{fla_repo} for all architectures.

\section{What does long-context attention require?}
\label{sec:balance}

In this section, we provide a framework for understanding what attributes of attention techniques make them suitable for long-context training. We focus on classic attention and linear attention, and conclude that neither is suitable for this setting. All experiments are conducted on LongCrawl64 \citep{lc64}, a dataset containing 6M documents each of length 64k tokens.

\subsection{Long-context attention requires a large state.}
\label{sec:a_large_state}

\begin{figure}
    \centering
    \begin{subfigure}[t]{0.248\textwidth}
        \includegraphics[width=\linewidth]{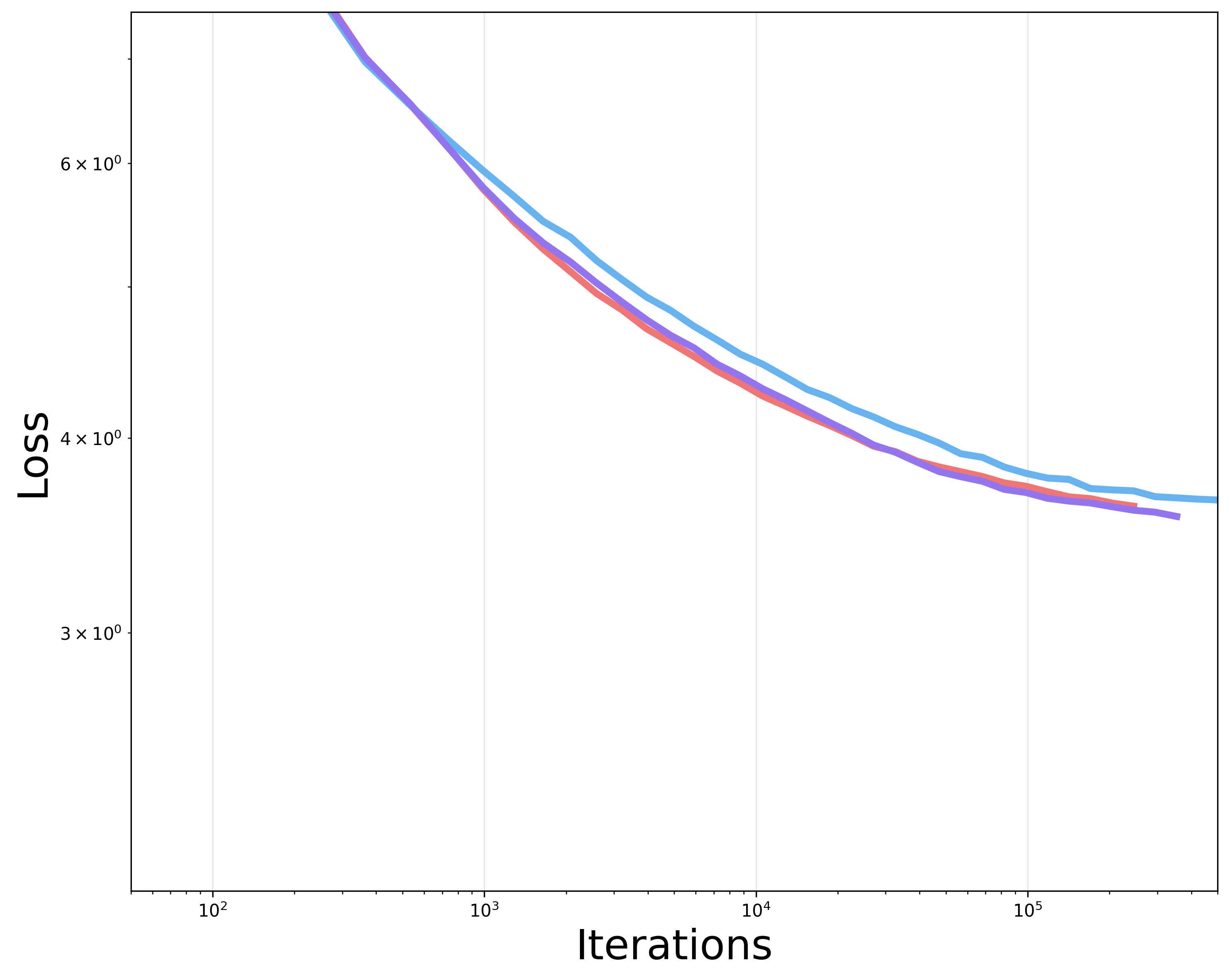}
        \caption{Context length 32.}
        \label{fig:lin_vs_exp_32}
    \end{subfigure}%
    \begin{subfigure}[t]{0.248\textwidth}
        \includegraphics[width=\linewidth]{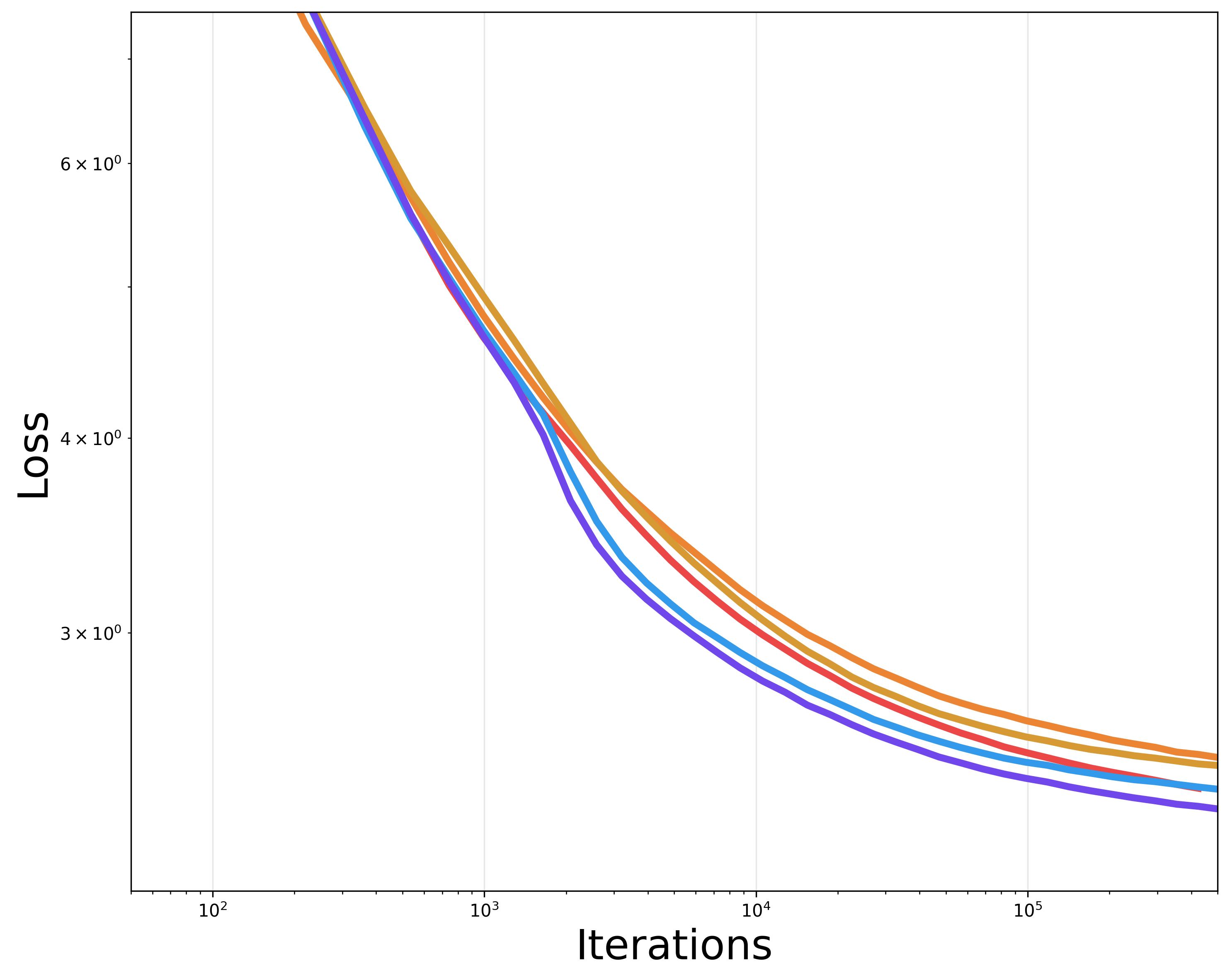}
        \caption{Context length 1024.}
        \label{fig:lin_vs_exp_1k}
    \end{subfigure}%
    \begin{subfigure}[t]{0.248\textwidth}
        \includegraphics[width=\linewidth]{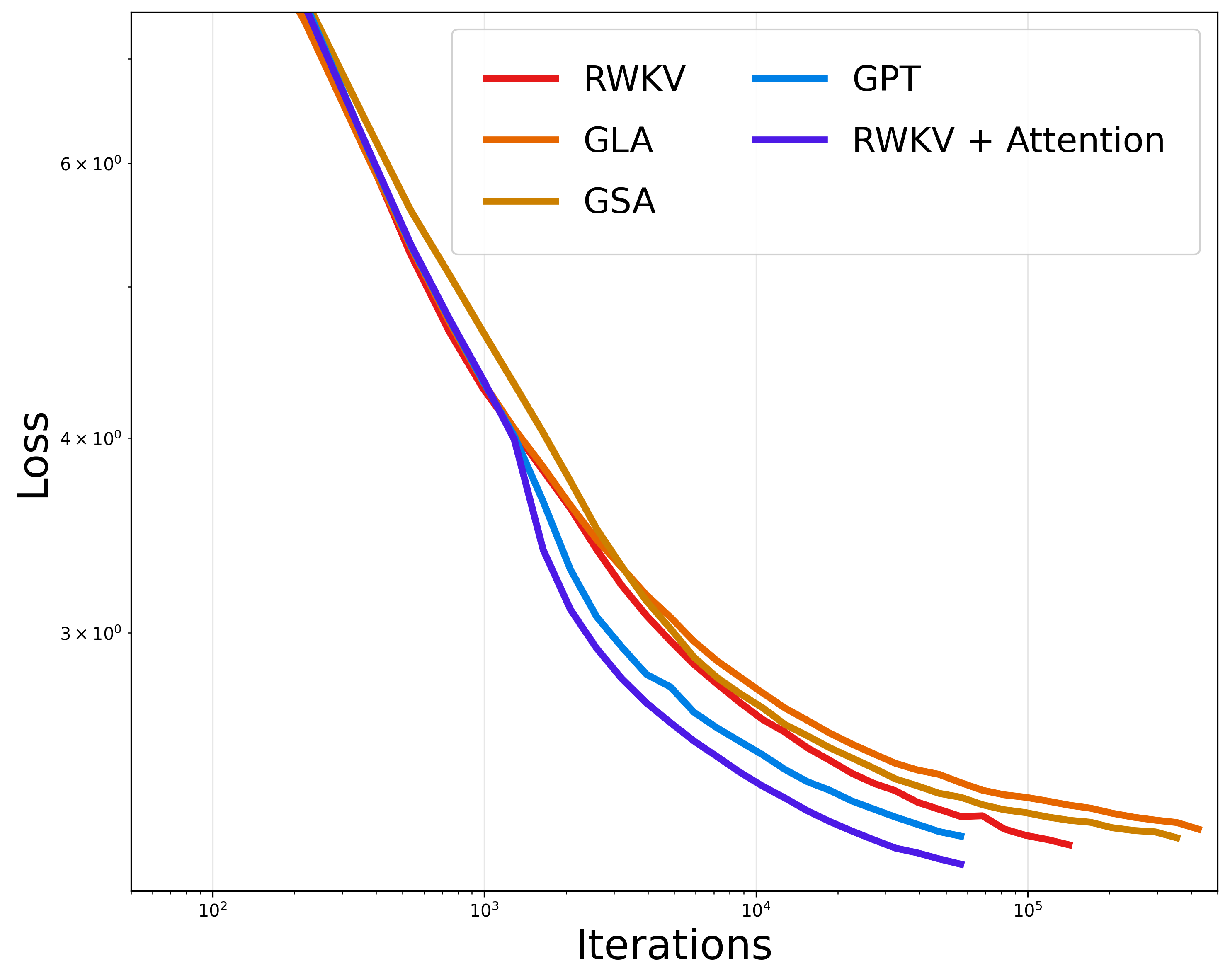}
        \caption{Context length 8192.}
        \label{fig:lin_vs_exp_8k}
    \end{subfigure}
    \begin{subfigure}[t]{0.248\textwidth}
        \includegraphics[width=\linewidth]{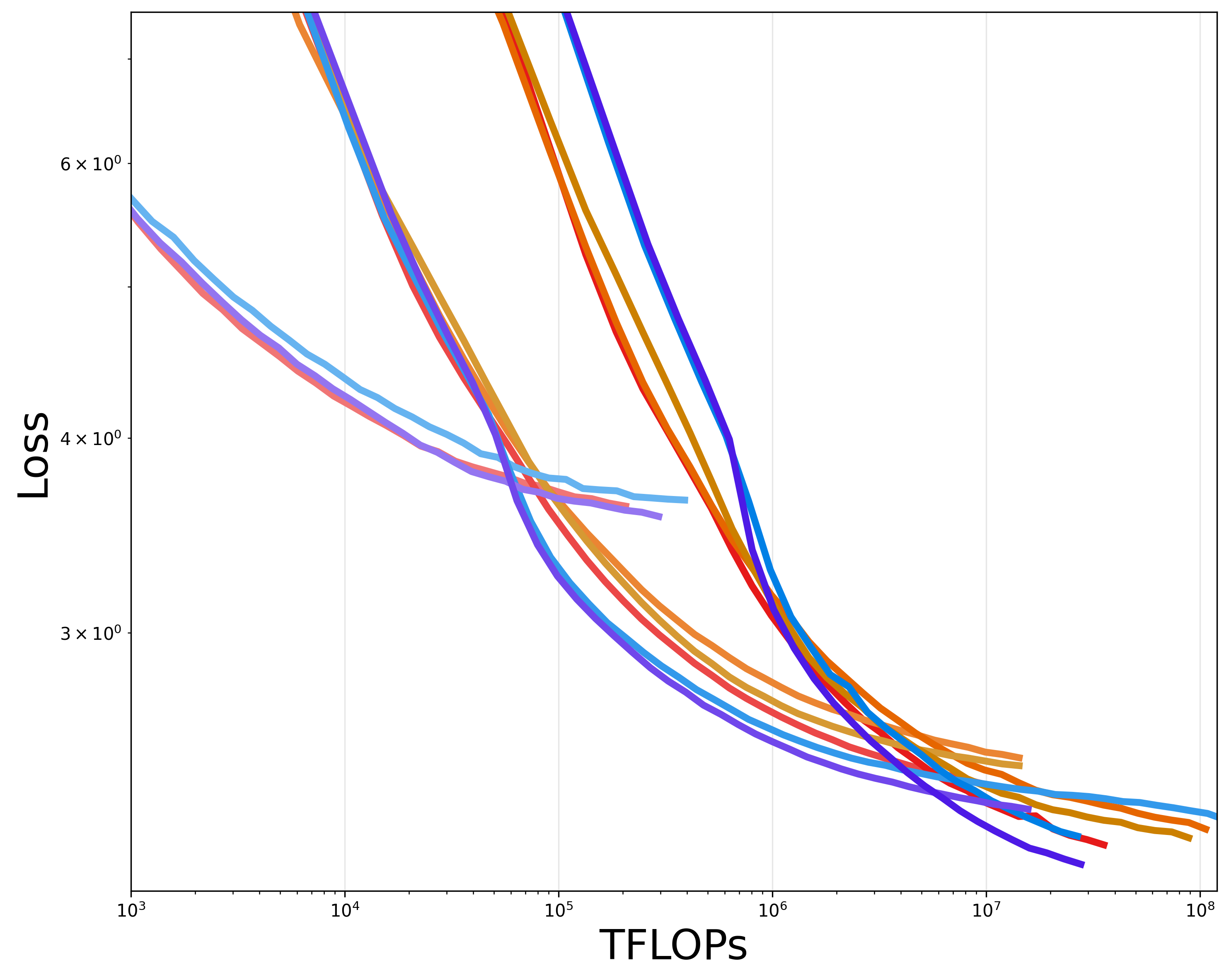}
        \caption{All curves, per-FLOP.}
        \label{fig:lin_vs_exp_all}
    \end{subfigure}%
    \caption{Exponential attention (blue) vs linear attention (red).}
    \label{fig:lin_vs_exp}
\end{figure}

In Figure~\ref{fig:lin_vs_exp} we compare the performance of classic exponential attention with that of linear attention at context lengths 32, 1024, and 8192. To check that our conclusions about attention are broadly applicable, we run these experiments across a range of architectures, each modified to use either exponential attention (blue) or linear attention (red). See Appendix~\ref{appendix:experimental_details} for the full experimental details.

At short context lengths, both forms of attention perform equivalently. But as context length grows, exponential attention gains an advantage.
We hypothesize a simple explanation for these observations: state scaling improves performance.
In this setting, at context length 32 (Figure~\ref{fig:lin_vs_exp_32}), the state size of linear and exponential attention is the same, explaining their equivalent performance per-update.
Whereas at context length 8k (Figure~\ref{fig:lin_vs_exp_8k}), the state size of exponential attention is 256x larger, explaining its better performance per-update.

This additional performance comes at a cost: the larger state requires additional FLOPs per update.
Linear attention is often claimed to be superior to exponential attention because it reduces this cost \citep{retnet}.
However, Figure~\ref{fig:lin_vs_exp_all} reveals that this is misleading, as the best performance for any FLOP budget can be achieved by training with exponential attention.

\subsection{Long-context attention requires state-weight balance.}
\label{sec:state_weight_balance}

We now explore the implications of the importance of state size on compute-optimal sequence architectures. The computations of a sequence model can be divided into \textit{weight FLOPs}, which involve an activation and a parameter, and \textit{state FLOPs}, which involve an activation and a state.\footnote{In this work, we only consider architectures whose weight FLOPs are proportional to parameter count, and whose state FLOPs are proportional to state size. However, note that techniques such as mixture-of-experts \citep{moe} produce a distinction between parameter count and weight FLOPs, and would require more nuanced analysis.} We have seen in the previous section that long-context performance scales with state size, and it is well-established that performance scales with parameter count \citep{scaling_law_for_lm}.

\begin{figure}[htbp]
  \centering
  \begin{minipage}[t]{0.48\textwidth}
    \vspace{0pt}\centering
    \includegraphics[width=\linewidth]{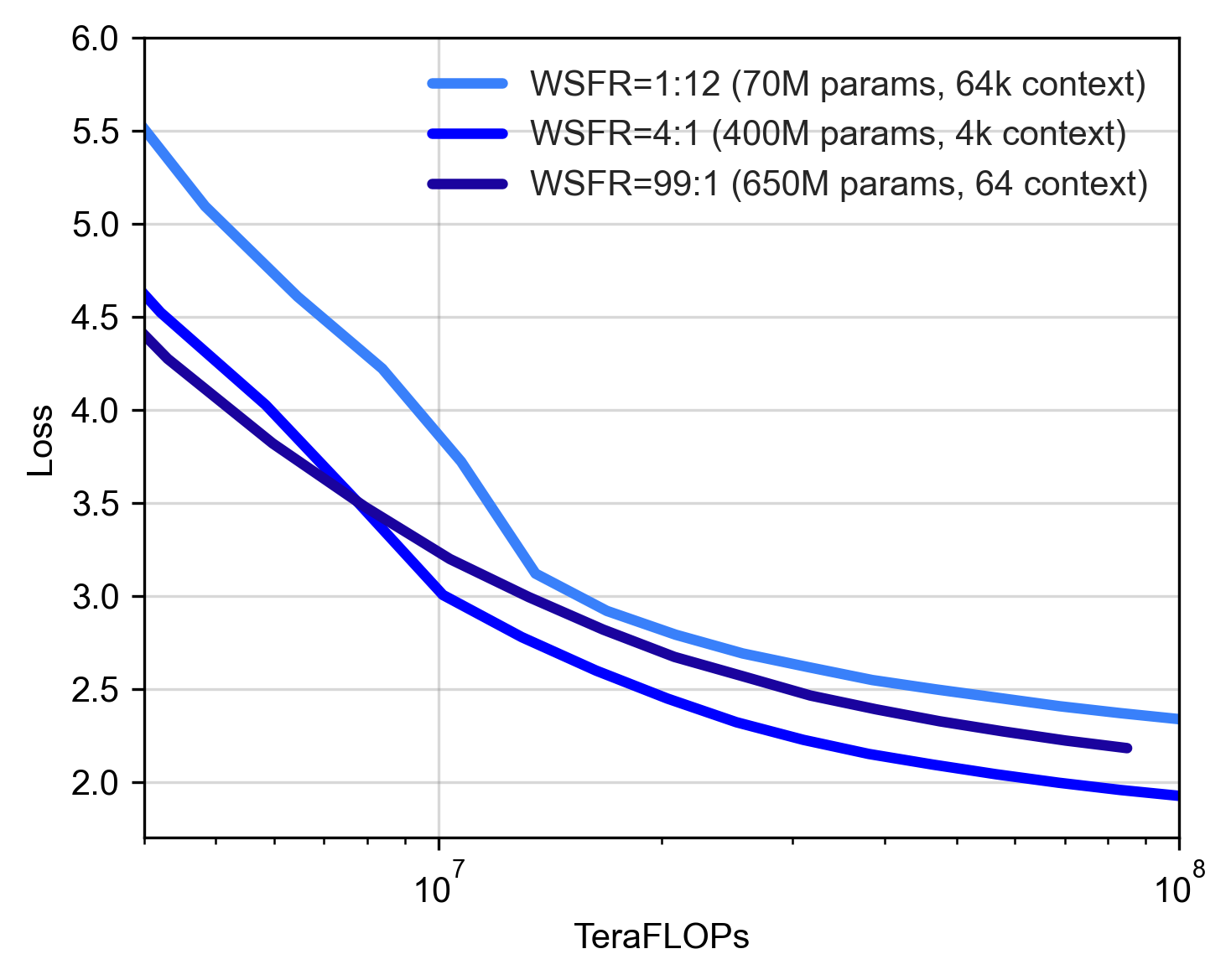}
    \caption{Compute-optimal transformers have a balanced WSFR. See Appendix~\ref{appendix:experimental_details} for details.}
    \label{fig:unbalanced_transformers}
  \end{minipage}\hfill
  \begin{minipage}[t]{0.48\textwidth}
    \vspace{0pt}\centering\small
    \begin{tabular}{lcc}
      \toprule
      \textbf{Attention} & \textbf{Context Length} & \textbf{WSFR} \\
      \midrule
      Exponential & 1\,024      & 8:1 \\
      \textbf{Exponential} & \textbf{8\,192}      & \textbf{1:1} \\
      Exponential & 65\,536     & 1:8 \\
      Exponential & 1\,000\,000 & 1:125 \\
      Linear      & 1\,024      & 30:1 \\
      Linear      & 8\,192      & 30:1 \\
      Linear      & 65\,536     & 30:1 \\
      Linear      & 1\,000\,000 & 30:1 \\
      Window-8192 & 1\,024      & 8:1 \\
      \textbf{Window-8192} & \textbf{8\,192}      & \textbf{1:1} \\
      \textbf{Window-8192} & \textbf{65\,536}     & \textbf{1:1} \\
      \textbf{Window-8192} & \textbf{1\,000\,000} & \textbf{1:1} \\
      \bottomrule
    \end{tabular}
    \captionof{table}{WSFR comparison between attention techniques at various context lengths. Balanced architectures are in bold.}
    \label{tab:compare_weight_state_flops}
  \end{minipage}
\end{figure}

We refer to as the relative proportion of these two types of FLOPs as the \textit{weight-state FLOP ratio (WSFR)}, and we argue that for compute-optimal models, the WSFR should be somewhat close to 1:1. This is because, for any model with a skewed WSFR (for example 100:1), doubling the smaller dimension will be effectively free in terms of total FLOPs. Since both the state and weight scales have a large impact on model performance, it is unwise to not take advantage of free scaling, and doing so will cause the WSFR to approach 1:1.

In Figure~\ref{fig:unbalanced_transformers}, we explore this empirically. We train a 400M GPT-2 model on context length 4096, as well as two other models with approximately the same total FLOPs: a small model with a large state, and a large model with a small state. \footnote{Since we adjust the state size via the context length, we also adjust the batch size inversely, to keep tokens-per-update identical between runs.} This set of models is therefore nearly identical except for WSFR, allowing us to isolate the impact of balance. We confirm that the most balanced architecture has the best performance.

Table~\ref{tab:compare_weight_state_flops} shows the WSFR of 124M-parameter GPT-2 models with various attention techniques and context lengths. Exponential attention is balanced for intermediate context lengths, but unbalanced for long context lengths, where it does far more state FLOPs than weight FLOPs. Linear attention, in contrast, is unbalanced at all context lengths in the opposite direction: far more weight FLOPs than state FLOPs. Thus, neither architecture is well-suited for long-context training.

How can we resolve this imbalance? One natural approach is to reduce the state size of exponential attention. In fact, many recent works in the transformer literature can be interpreted through this lens: hybrid architectures \citep{jamba} reduce the size of the state along the layer dimension, sparse attention \citep{sparse_transformers} reduces the size of the state along the time dimension, multi-query attention \citep{mqa} reduces the size of the state along the head dimension, and latent attention \citep{mla} reduces the size of the state along the feature dimension. We use windowed attention to exemplify this family of \textit{reduced-state exponential attention} approaches. Table~\ref{tab:compare_weight_state_flops} shows that windowed attention architectures have balanced WSFR for large context lengths, given appropriate selection of window size.

\subsection{Long-context attention requires in-context learning.}

\begin{figure}
  \centering
  \begin{minipage}[t]{0.38\textwidth}
    \vspace{0pt}
    \centering
    \includegraphics[width=\linewidth]{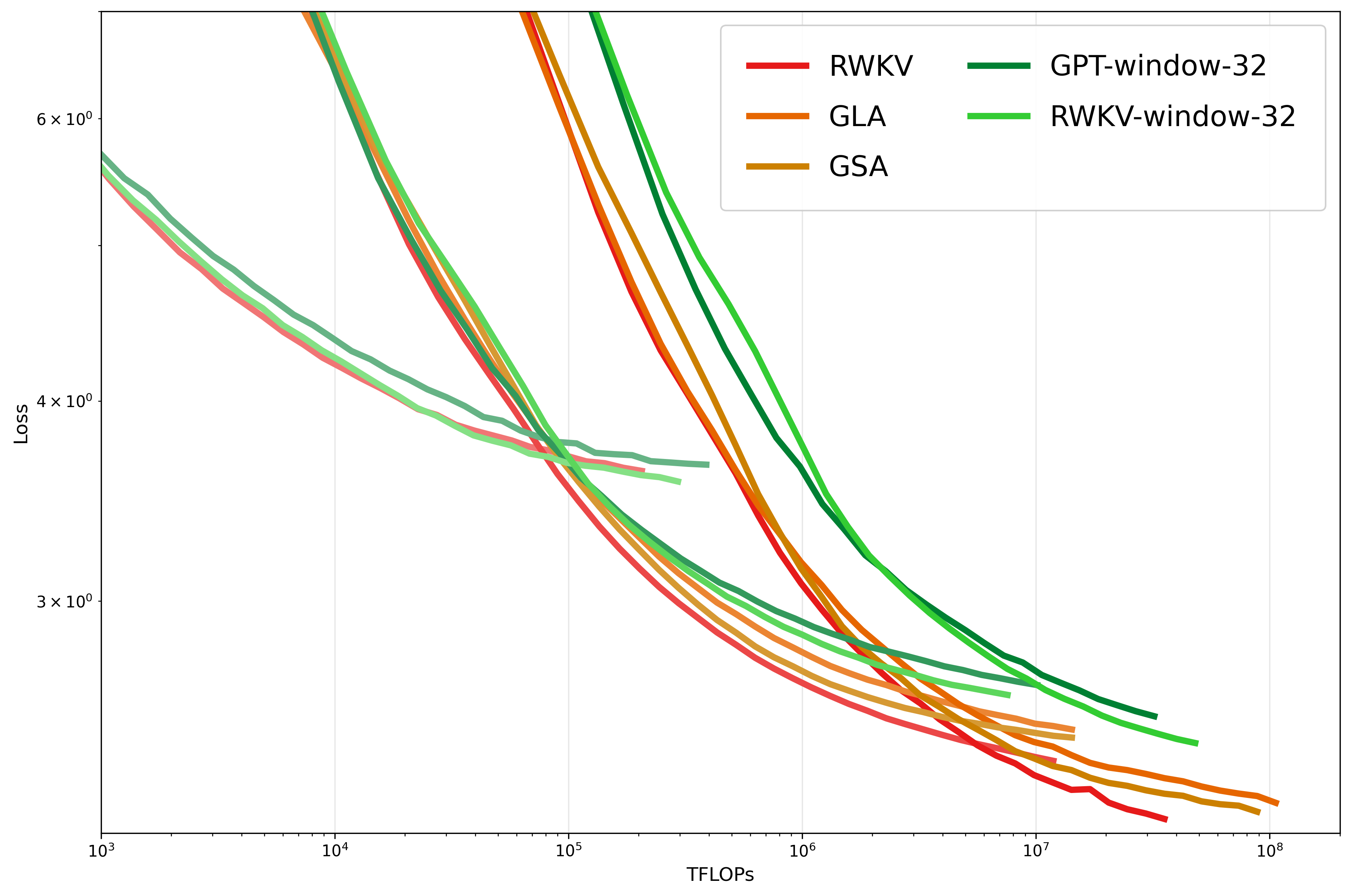}
    \captionof{figure}{Comparing learning curves if linear vs window-32 attention for several architectures and context lengths.}
    \label{fig:icl_comparison_combined}
  \end{minipage}
  \hfill
  \begin{minipage}[t]{0.58\textwidth}
    \vspace{0pt}
    \centering
    \begin{subfigure}[t]{0.48\textwidth}
      \includegraphics[width=\linewidth]{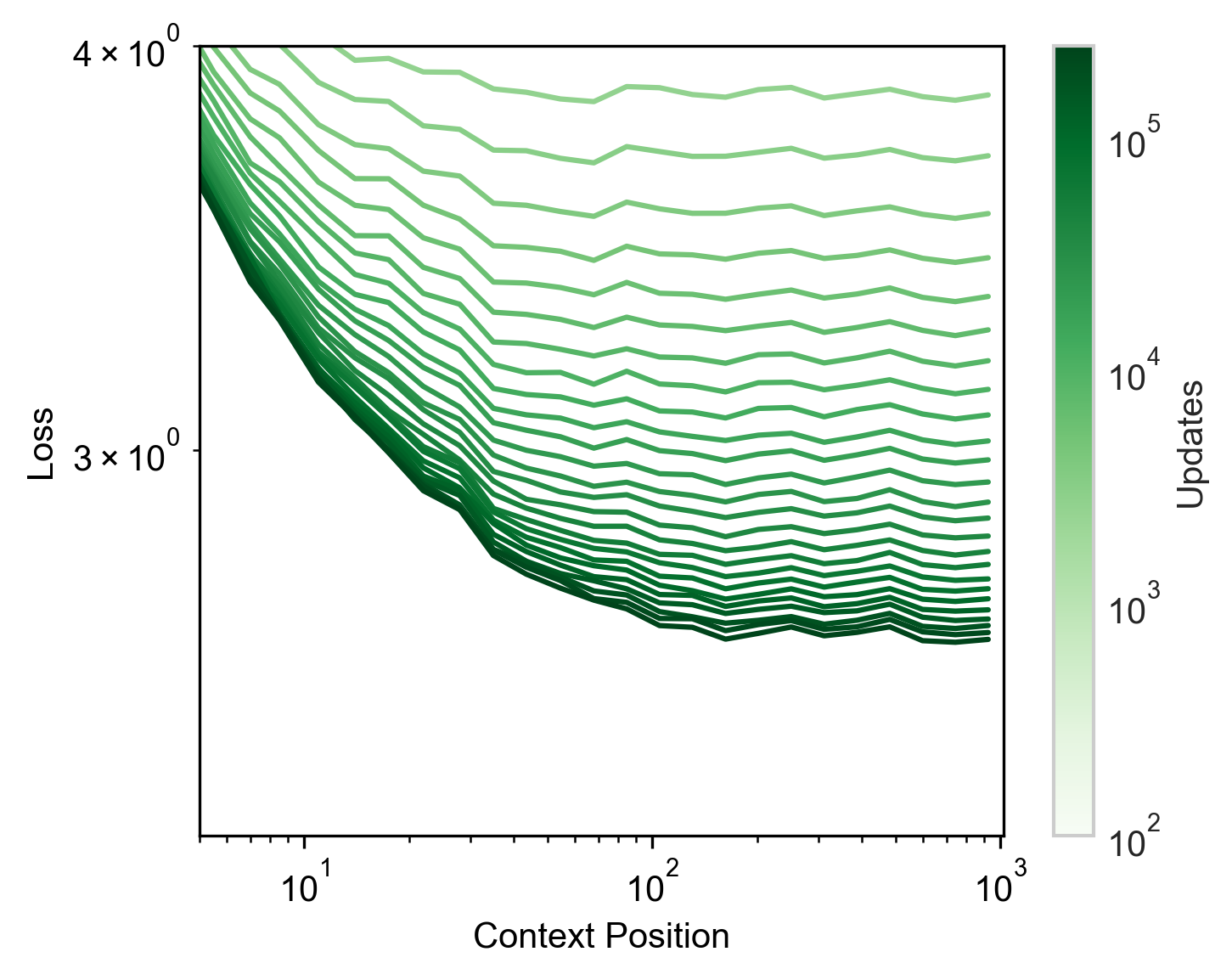}
      \caption{Window-32 attention.}
      \label{fig:icl_comparison_w32}
    \end{subfigure}\hfill
    \begin{subfigure}[t]{0.48\textwidth}
      \includegraphics[width=\linewidth]{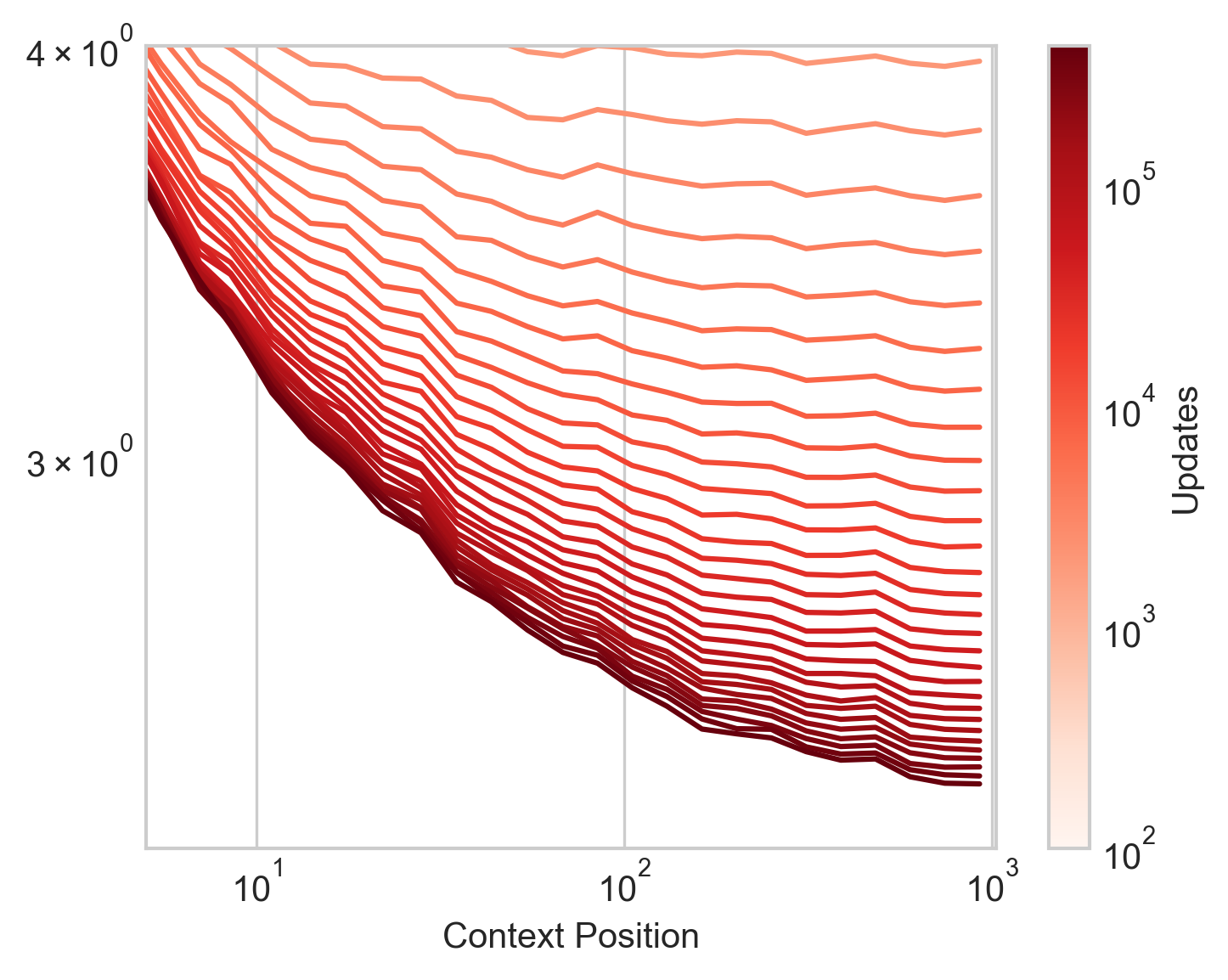}
      \caption{Linear attention.}
      \label{fig:icl_comparison_rwkv}
    \end{subfigure}
    \captionof{figure}{In-context learning across training for RWKV variants at context length 1024. }
    \label{fig:icl_comparison}
  \end{minipage}\hfill
\end{figure}

In Section~\ref{sec:a_large_state}, we saw exponential attention outperform linear attention, and attributed this success to its larger state. In Figure~\ref{fig:icl_comparison_combined}, we perform a more fair comparison, by juxtaposing linear attention with windowed attention of equal state size. See Appendix~\ref{appendix:experimental_details} for details. We see a reversal of the previous trend: it is now linear attention that dominates at all context lengths and FLOP budgets. This indicates that linear attention makes better use of its state than windowed attention.

We can explain this gap using an \textit{in-context learning (ICL) curve} of the training loss, which plots the negative log-likelihood at each context length throughout training. In Figure~\ref{fig:icl_comparison}, we compare the in-context learning ability of RWKV with windowed attention to linear attention up to context length 1024. Figure~\ref{fig:icl_comparison_w32} shows that no in-context learning occurs beyond 100 tokens for window-32 attention.\footnote{Note that this is a 12-layer model, so the effective context window is $12*32=384$ tokens.} In contrast, linear attention can be seen in Figure~\ref{fig:icl_comparison_rwkv} to demonstrate consistent in-context learning across the entire sequence.

Ultimately, a long-context model only has value if the extra context improves its predictions, so these results tell us that windowed attention, despite being balanced at all context lengths, is nonetheless a poor choice for a long-context architecture. We hypothesize that this limitation will be shared by the other reduced-state exponential attention approaches discussed in Section~\ref{sec:state_weight_balance}, although thorough investigation of this hypothesis is left to future work. Instead, in Section~\ref{sec:powatt}, we introduce a technique for the other natural approach to balanced long-context attention, \textit{expanded-state linear attention}.

\section{Power attention}
\label{sec:powatt}

If one substitutes the exponential in the classic attention formula by the $p$-th power, the result is \textit{power attention}, variants of which have been studied by \citet{based, polysketch}.
\begin{align}
\powattn(Q,K,V)_i =  \sum_{j=1}^i  \left ( Q^T_i K_j \right )^p V_j
\end{align}
Power attention is a special case of linear attention because there exist functions $\phi$ s.t. $  \phi(Q_i) ^T \phi(K_j) = \left ( Q^T_i K_j \right )^p$, granting the computational advantages of linear attention discussed in Section~\ref{sec:background}. Its simple inner-product attention form gives it an important computational advantage over other proposed state-expanded linear transformers, such as DPFP described in \cite{linear_transformer_fast_weight_programmers}, which require the explicit expansion of $\phi(q), \phi(k)$ in the attention form. When large intermediate objects (such as expanded keys) are involved, the fused attention algorithms pioneered by \cite{flash2} do not work, meaning such algorithms have poor hardware utilization in practice.

\begin{lemma}
The function $\tpow_p: \R^d \to \R^{d^p}$ defined as
\begin{align}
\tpow_p(x) =  
\begin{bmatrix}
x_1 \cdots x_1 \\
 x_1 \cdots x_2 \\
\vdots \\
 x_d \cdots x_d \\
\end{bmatrix}
=
\begin{bmatrix}
\vdots \\
 \prod_k x_{i_k} \\
\vdots \\
\end{bmatrix}_{(i_1, \cdots, i_p)\in \mi}
\end{align}
Then, for $q,k\in \R^d$ the following property holds $\tpow_p(q)^T \tpow_p(k) = (q^T k)^p$
\label{lemma:tpow_inner_product}
\end{lemma}
We therefore have that $ \powattn(Q,K,V) = \linattntpow(Q,K,V)$. The proofs for this section can be found in \autoref{sec:derivation_of_spow}.

However, a major disadvantage of $\tpow_p(x)$ is that it contains redundant entries. The theory of \textit{symmetric powers} can be used to address this issue.

\begin{lemma}
For any $d,p\in\N$ denote the set of \textit{non-decreasing multi-indices} as $\ndmi = \{ (i_1, \cdots, i_p) \in \mi \; | \; i_1 \le \cdots \le i_p \}$. Define $\spow_p: \R^d \to \R^D$ to be the function
\begin{align}
\spow_p(x)
=
\begin{bmatrix}
\vdots \\
\sqrt{\frac{p!}{\prod_k \hist_k(i)!}}   \;  \prod_k x_{i_k} \\
\vdots \\
\end{bmatrix}_{i\in \ndmi}
\end{align}
Where $\hist_k(i_1,\cdots,i_2) = \sum_{j=1}^p 1(i_j = k)$ is simply the function that counts how many times the index $k$ occurs across the the multi index. Then, the following statements hold:
\begin{enumerate}
    \item The dimensionality $D$ is given by $\binom{d + p - 1}{p}$   (the binomial n choose k)
    \item The inner products $\spow_p(q)^T \spow_p(k) = (q^T k)^p$
\end{enumerate}
\label{lemma:spow_inner_product}
\end{lemma}
A few concrete examples might be helpful:
$$
\spow_2 \left(
\begin{bmatrix}
x_1 \\
x_2
\end{bmatrix}
\right)
=
\begin{bmatrix}
  \;\;\;\;\;\; x_1 x_1 \\
  \sqrt{2} \; x_1 x_2 \\
  \;\;\;\;\;\; x_2 x_2 \\
\end{bmatrix}
\qquad 
\spow_3 \left(
\begin{bmatrix}
x_1 \\
x_2
\end{bmatrix}
\right)
=
\begin{bmatrix}
 \;\;\;\;\;\;  x_1 x_1  x_1\\
  \sqrt{3} \; x_1 x_1 x_2 \\
  \sqrt{3}  \; x_1 x_2 x_2 \\
  \;\;\;\;\;\; x_2 x_2 x_2 \\
\end{bmatrix}
$$
Table \ref{tab:savings} compares the dimensions $d^p$ and $\binom{d + p - 1}{p}$ for $\tpow$ and $\spow$ respectively. For large $p$, these reductions in $D$ have a large impact on the runtime and memory utilization of chunked power attention.

Ultimately, $\spow_p$ is a state expansion that increases the state size by a factor of $\frac{\binom{d + p - 1}{p}}{d}$ without introducing any parameters. For example, for a model with head size 64, $p=2$ increases the state size by a factor of approximately 32, $p=3$ by about 700, and $p=4$ by about 12000.

\begin{wraptable}{r}{0.5\textwidth}
\centering
\begin{tabular}{c|c|c|c}
$p$ & $\tpow \; D$ & $\spow\; D$ & Savings \\
\hline
2 & 4096 & 2080 & 49\%\\
3 & 262144 & 45760 & 82\%\\
4 & 16777216 & 766480 & 95\%\\
5 & 1073741824 & 10424128 & 99\%\\
6 & 68719476736 & 119877472 & 99.8\%\\
\end{tabular}
\caption{State size comparison for tensor vs.\ symmetric power embeddings for $d=64$.}
\label{tab:savings}
\end{wraptable}


\subsection{Hardware-aware implementation}
\label{sec:kernels}

An efficient implementation of chunked power attention requires careful consideration to the sizes of relevant objects. The main quantities that appear are the key dimension $d$, the value dimension $v$, the sequence dimension $t$, and the expanded key dimension $D$. At large problem sizes, $d,v$ typically stay small, whereas $t,D$ typically become large. For example, in Llama 3 \citep{llama3}, the largest $d$ is $128$, whereas the largest $t$ is $128000$.

The inputs and outputs of attention,  $Q,K,V,Y$, are all in either $\R^{t \times d}$ or $\R^{t \times v}$. But some intermediate objects, most notably $\phi(Q), \phi(K)$, live in $\R^{t \times D}$. If materialized, these objects dominate memory consumption, and their IO bottlenecks computation and reduces arithmetic intensity.
This is reminiscent of how in standard attention, memory and IO is dominated by the attention matrix, an intermediate object in $\R^{t \times t}$. This problem was addressed by Flash Attention \citep{flash} via operator fusion, whose central algorithmic innovation was the design of a kernel that avoids materializing the attention matrix in HBM.

We apply the same principles to design efficient kernels for chunked power attention. We factorize the algorithm as follows:
\begin{align}
\text{update-state}(S, K, V) = S + V^T \phi(K) \qquad \text{query-state}(S, Q) =  \phi(Q) S^T
\end{align}
Each of these functions centers around a matrix multiplication between a large expanded object and a smaller object. This computational structure can be exploited via a fused \textit{expand-MMA} kernel, a matrix multiplication where the tiles of one operand are expanded on-the-fly.

Our implementation of expand-MMA uses Triton \citep{triton} with a custom templating system to handle multiple values of $p$.
A complete implementation of chunked power attention also requires a kernel for intra-chunk attention and another for the cumulative gated sum of states (see Appendix \ref{sec:algorithms}). We use Flash Attention \citep{flash} for the former and a simple CUDA kernel for the latter.

We have released our kernels open-source\footnote{\url{https://github.com/m-a-n-i-f-e-s-t/power-attention}} to allow others to use power attention, and to enable research on other applications of the symmetric power in deep learning. In Appendix \ref{sec:algorithms}, we provide more details on our implementation.

\subsubsection{$\tiledspow$}
Based on Table \ref{tab:savings}, one would expect that chunked linear attention using $\spow$ would run faster than $\tpow$, because its smaller $D$ translates into fewer FLOPs.
However, modern GPUs are mainly optimized for matrix multiplications, and the $\tpow$ expansion is more compatible with the computational structure of a matmul.
$\tpow$ calculations can be easily partitioned for parallel processing amongst CTAs/threads using a standard tiling approach. In contrast, the less-regular structure of a symmetric tensor is less compatible with standard GPU operations. The correction term $\sqrt{\frac{p!}{\prod_k \hist_k(i)!}}$ must be computed on slower CUDA cores, causing thread divergence due to branching, and the jagged memory access patterns lead to share memory bank conflicts.

\begin{wrapfigure}{r}{0.5\textwidth}
\centering
\begin{subfigure}[b]{0.32\linewidth}
\centering
\begin{tikzpicture}
  \def\N{6} \def\size{0.3}
  \foreach \i in {0,...,5}\foreach \j in {0,...,5}{
    \definecolor{cellcolor}{rgb}{0.7,0.85,1}
    \fill[cellcolor] (\j*\size,-\i*\size) rectangle ++(\size,-\size);
    \draw (\j*\size,-\i*\size) rectangle ++(\size,-\size);
  }
  \draw[line width=.5pt] (0,0) rectangle (\N*\size,-\N*\size);
\end{tikzpicture}
\caption{$\tpow$}
\end{subfigure}
\hfill
\begin{subfigure}[b]{0.32\linewidth}
\centering
\begin{tikzpicture}
  \def\N{6} \def\size{0.3}
  \foreach \i in {0,...,5}\foreach \j in {0,...,5}{
    \ifnum\j<\i
      \draw (\j*\size,-\i*\size) rectangle ++(\size,-\size);
    \else
      \definecolor{cellcolor}{rgb}{0.7,0.85,1}
      \fill[cellcolor] (\j*\size,-\i*\size) rectangle ++(\size,-\size);
      \draw (\j*\size,-\i*\size) rectangle ++(\size,-\size);
    \fi
  }
  \draw[line width=.5pt] (0,0) rectangle (\N*\size,-\N*\size);
\end{tikzpicture}
\caption{$\spow$}
\end{subfigure}
\hfill
\begin{subfigure}[b]{0.32\linewidth}
\centering
\begin{tikzpicture}
  \def\N{6} \def\s{0.3}
  \foreach \R in {0,...,2}\foreach \C in {0,...,2}{
    \ifnum\C<\R\relax\else
      \definecolor{cellcolor}{rgb}{0.7,0.85,1}
      \fill[cellcolor] (2*\C*\s,-2*\R*\s) rectangle ++(2*\s,-2*\s);
    \fi
  }
  \foreach \i in {0,...,5}\foreach \j in {0,...,5}
    \draw (\j*\s,-\i*\s) rectangle ++(\s,-\s);
  \foreach \R in {0,...,2}\foreach \C in {0,...,2}{
    \ifnum\C<\R\relax\else
      \draw[line width=1pt,rounded corners=1pt]
            (2*\C*\s,-2*\R*\s) rectangle ++(2*\s,-2*\s);
    \fi
  }
  \draw[line width=.4pt] (0,0) rectangle (\N*\s,-\N*\s);
\end{tikzpicture}
\caption{$\tiledspow$}
\end{subfigure}
\caption{Illustration of $\tpow$, $\spow$, and $\tiledspow$.}
\label{fig:tiled_spow}
\end{wrapfigure}

Our approach is to use the idea of tiling to interpolate between $\tpow$ and $\spow$, harnessing benefits of both. Our proposed \textit{tiled symmetric power} expansion, $\tiledspow$, operates on tiles of data (providing the GPU-friendly structure of $\tpow$) but only computes tiles of data with non-decreasing multi-indices (reducing data duplication like $\spow$). Figure~\ref{fig:tiled_spow} paints the basic picture for $p=2$. The dimension of every tile is $\dtile^p$, and the number of tiles with non-decreasing multi indices is $\binom{d/\dtile + p - 1}{p}$. This means the dimension $D=\binom{d/\dtile + p - 1}{p} \dtile^p$. Empirically, we find that $\dtile=8$ is a good choice for $p=2$. For $p=3$ a smaller $\dtile=4$ seems preferable.

\begin{figure}
  \centering
  \begin{subfigure}[b]{0.3\textwidth}
    \includegraphics[width=\textwidth]{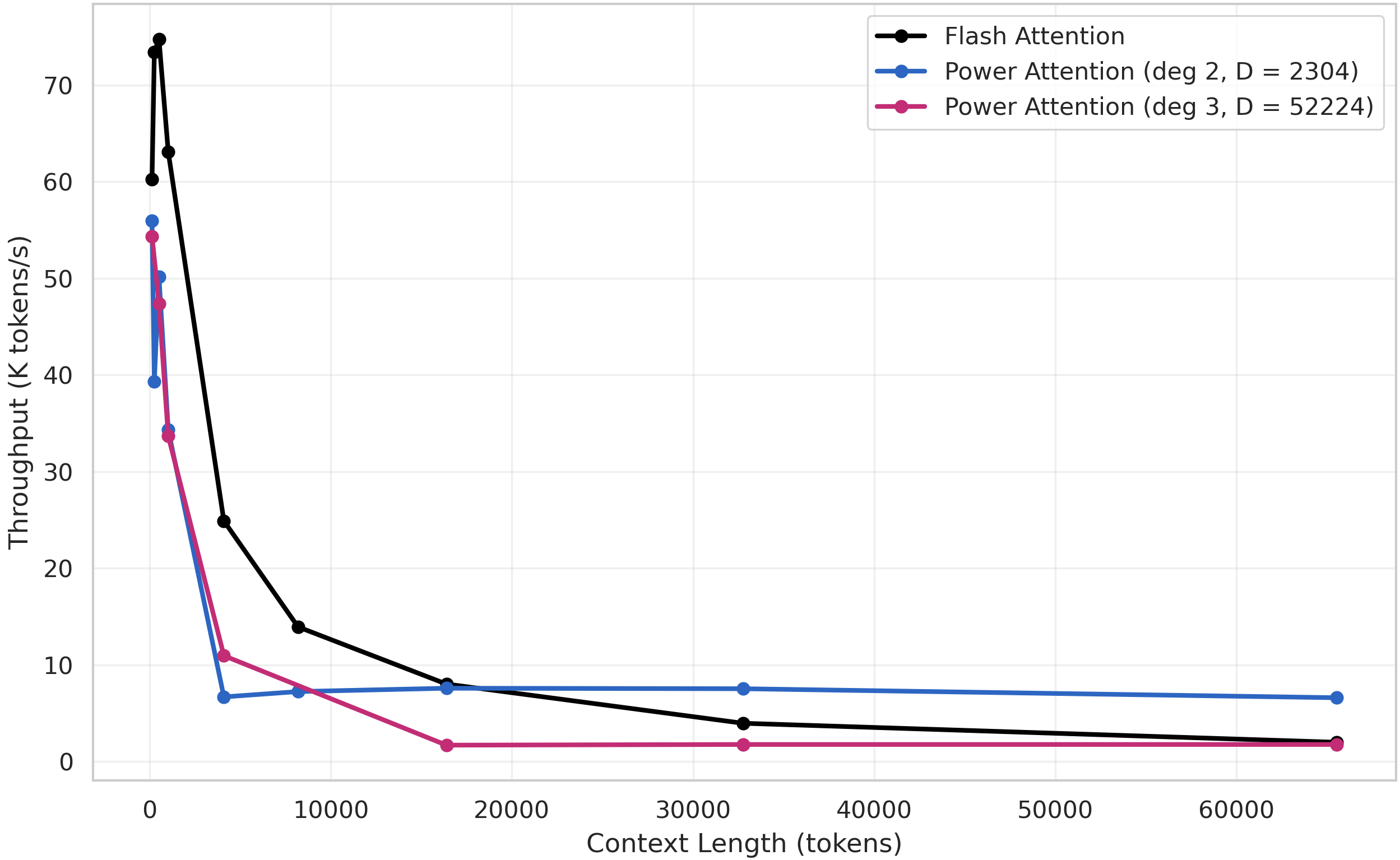}
    \caption{Throughput comparison at head size 64.}
  \end{subfigure}
  \hfill
  \begin{subfigure}[b]{0.3\textwidth}
    \includegraphics[width=\textwidth]{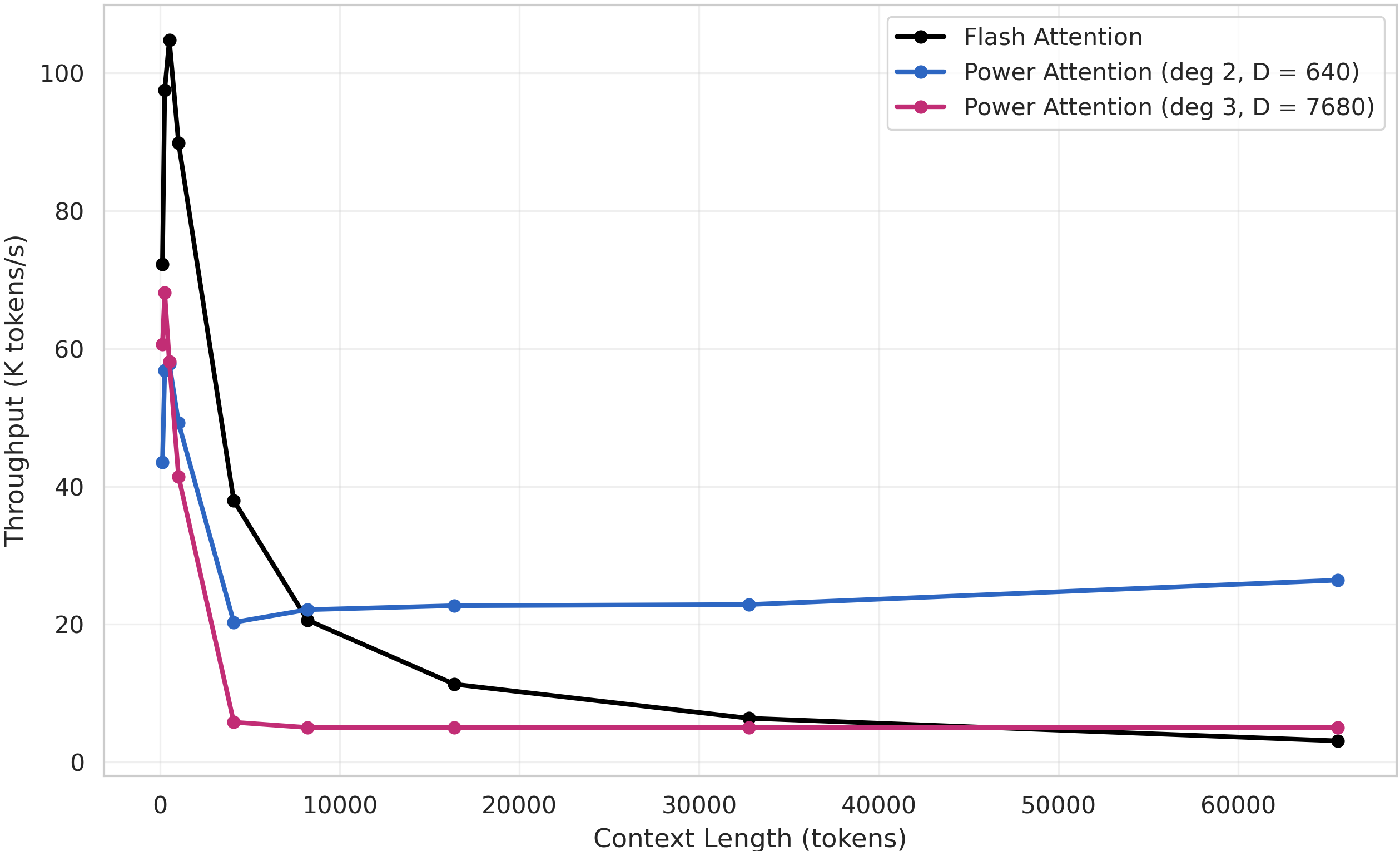}
    \caption{Throughput comparison at head size 32.}
  \end{subfigure}
\hfill
  \begin{subfigure}[b]{0.3\textwidth}
    \includegraphics[width=\textwidth]{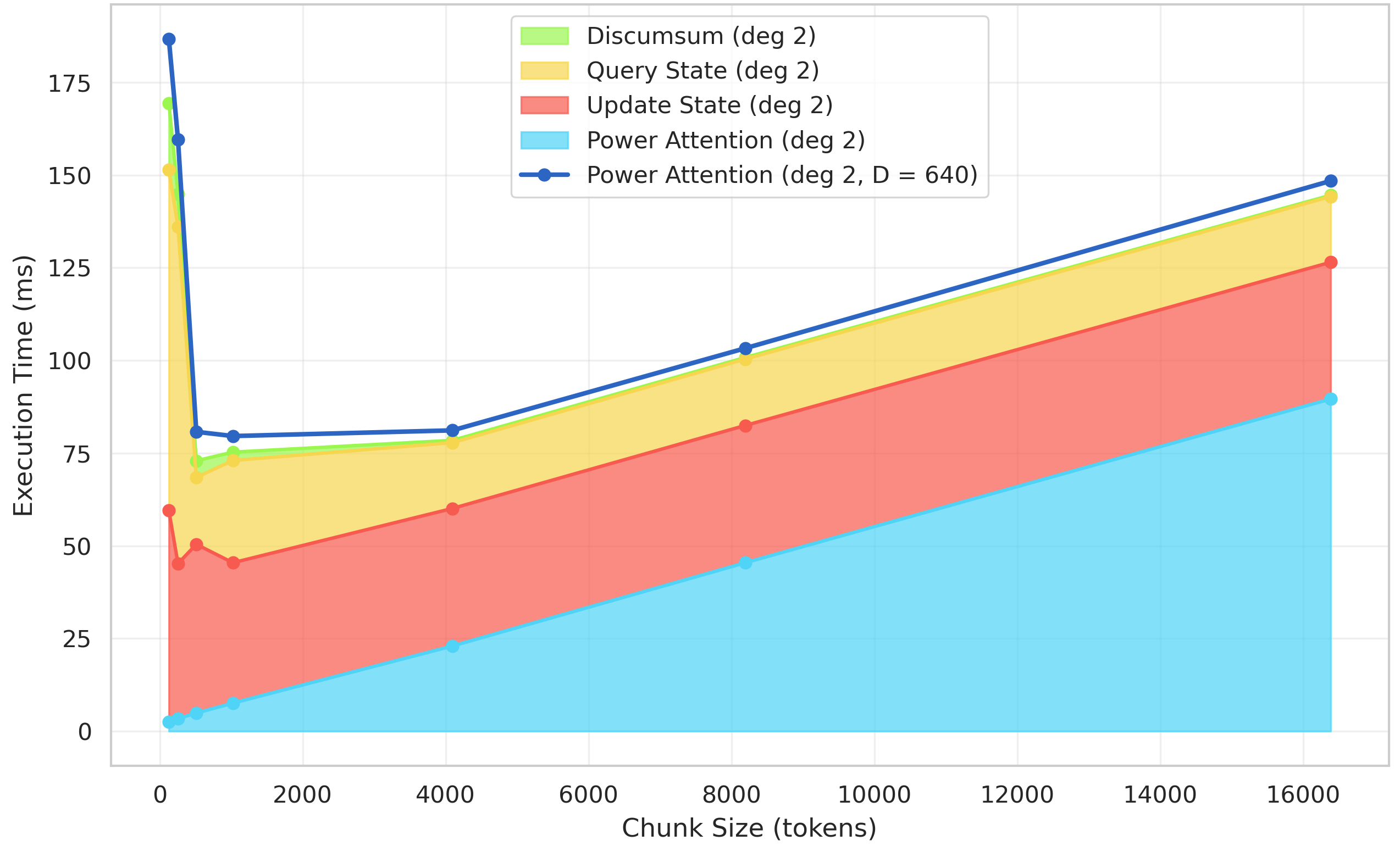}
    \caption{Effect of chunk size on total execution time.}
  \end{subfigure}
  \caption{Hardware efficiency of Power Attention kernels.}
\label{fig:benchmark_result}
\end{figure}

\subsubsection{Benchmarks}

To benchmark our progress, we compare the throughput (tokens per second) between Power Attention and Flash Attention kernels, with a batch size of 8 and 12 heads, on an A100 GPU. For short contexts, the attention form achieves higher throughput, but as the context size grows, Power Attention switches to the chunk form and retains a constant throughput from then on. On the other hand, the throughput of flash attention decays proportional to $t$. At context length 65536, degree-2 Power Attention achieves \textbf{3.3x} (for head size 64) and \textbf{8.6x} (for head size 32) higher throughput than Flash Attention.

Note that the performance of Power Attention is highly dependent on the chunk size $c$. Figure~\ref{fig:benchmark_result} shows the total execution time is broken down into its component operations for various $c$.

\section{Empirical evaluation of power attention}
\label{sec:experiments}

In this section, we evaluate power attention on the basis of its in-context learning ability and long-context performance. To ensure that the dataset contains documents with true long-term structure \footnote{This is not true of many common benchmark datasets. For example, most sequences in OpenWebText \citep{openwebtext} have length less than 1k. Figure~\ref{fig:long_data} in Appendix~\ref{appendix:longcrawl} shows the document length distribution.}, all of our experiments are conducted on LongCrawl64 \citep{lc64}. For these experiments, we use power attention with per-head gating, and normalize by the sum of the attention weights.

\subsection{In-context learning comparison}
\label{sec:experiments_icl}

\begin{figure}
    \centering
    \begin{subfigure}[t]{0.32\textwidth}
        \includegraphics[width=\linewidth]{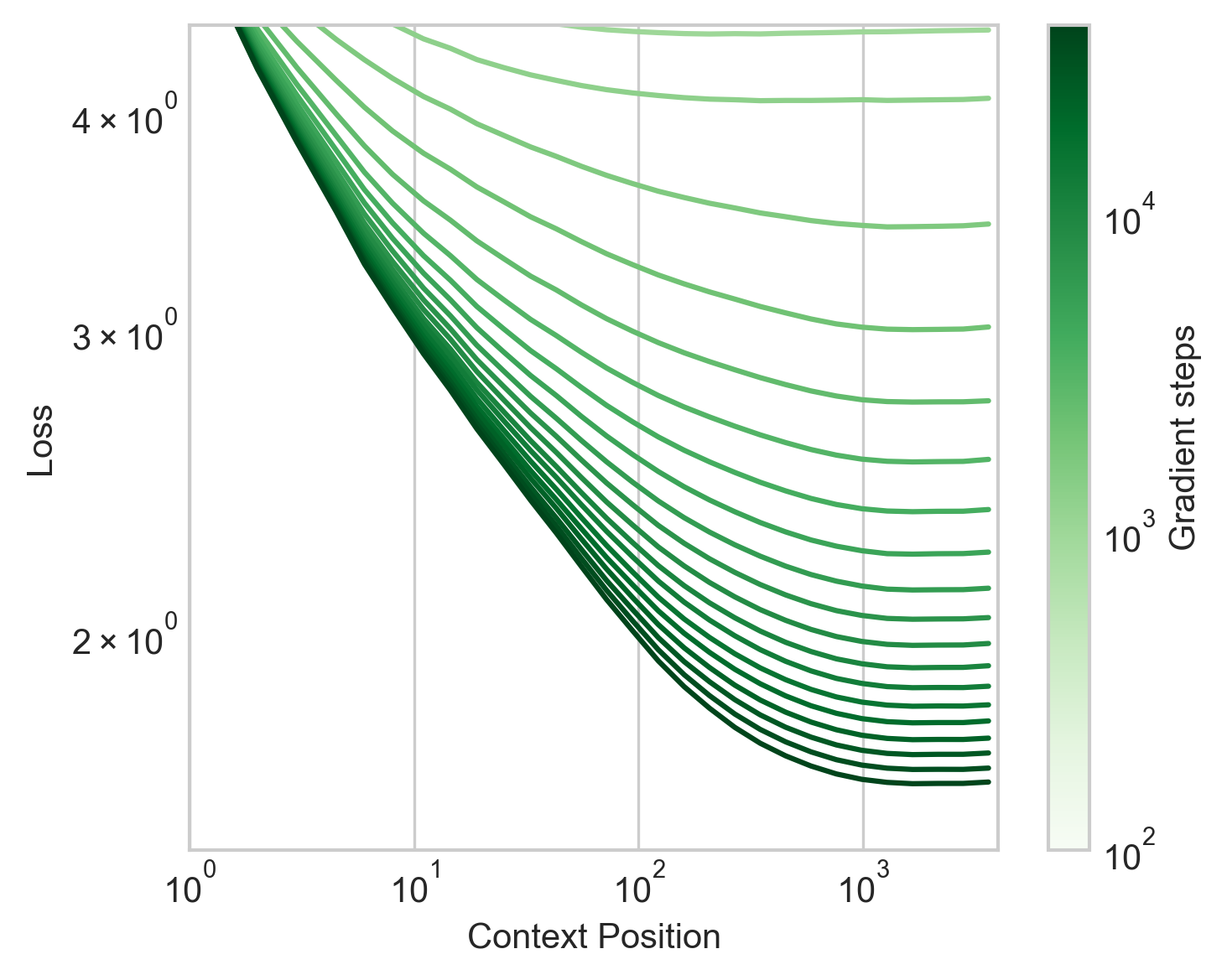}
        \caption{Window-1k attention.}
        \label{fig:yolo_attn1k}
    \end{subfigure}
    \begin{subfigure}[t]{0.32\textwidth}
        \includegraphics[width=\linewidth]{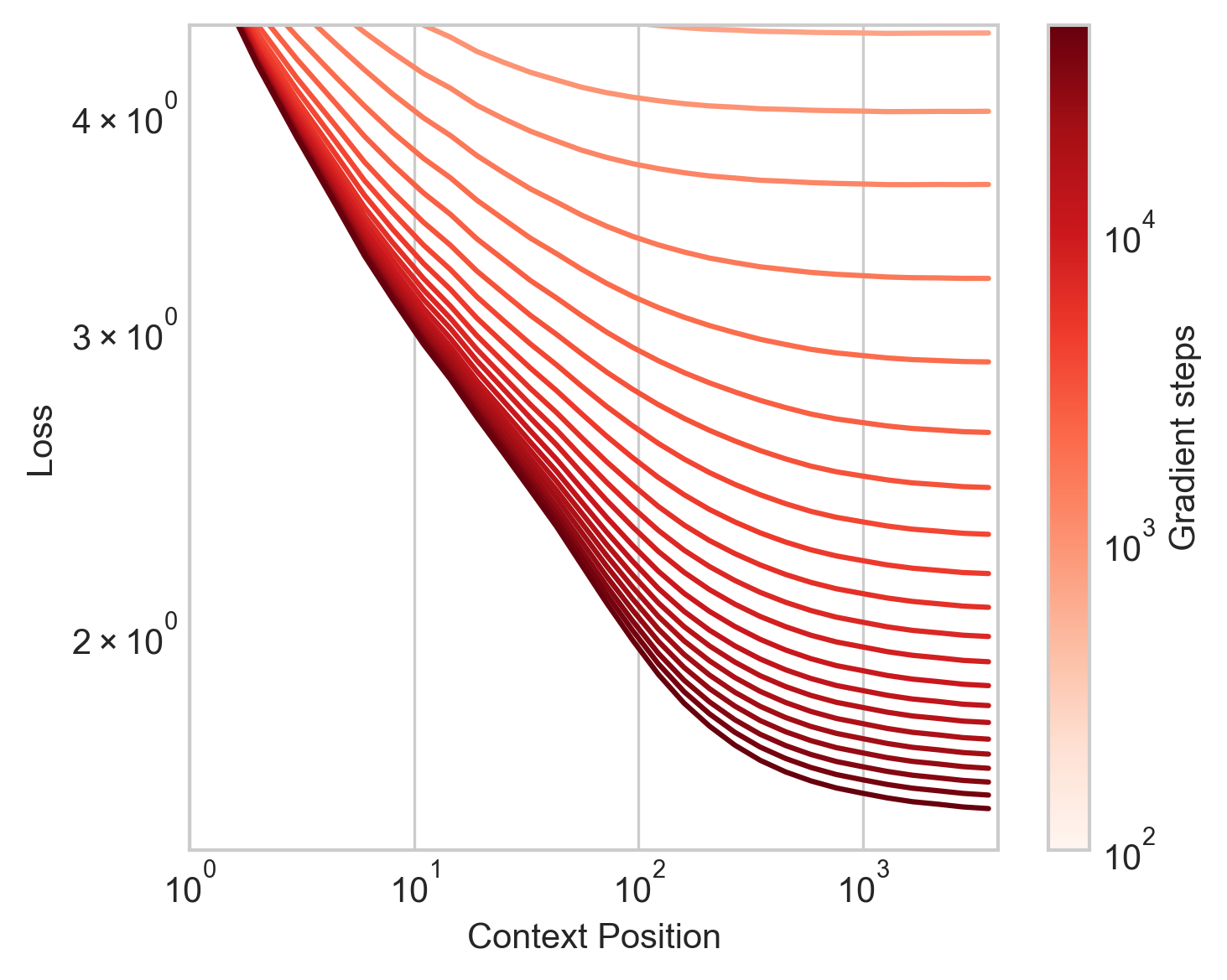}
        \caption{$p=2$ power attention.}
        \label{fig:yolo_p2}
    \end{subfigure}%
    \begin{subfigure}[t]{0.32\textwidth}
        \includegraphics[width=\linewidth]{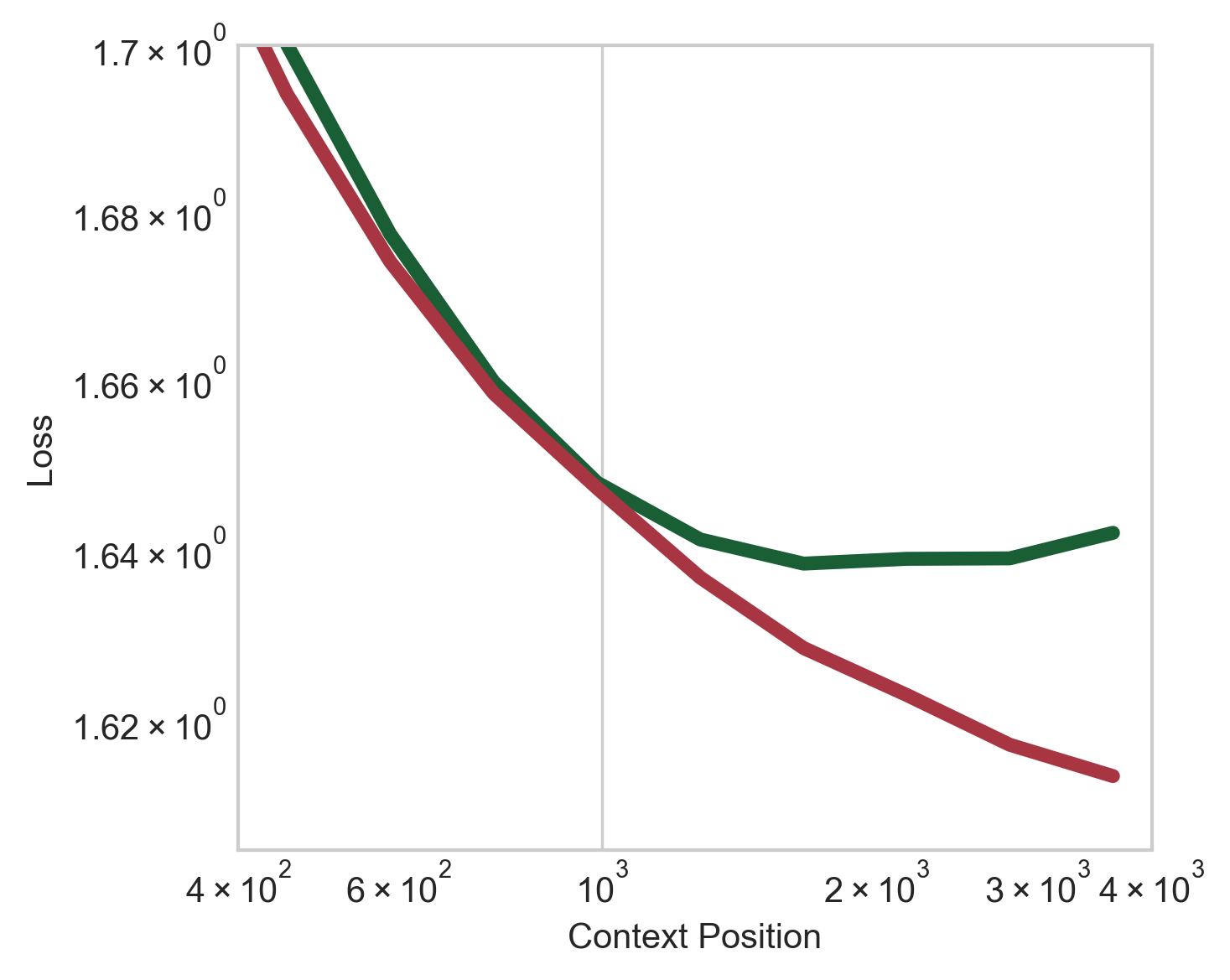}
        \caption{Close-up on ICL curves at 50k.}
        \label{fig:yolo_combined_curves}
    \end{subfigure}%
    \caption{Power attention demonstrates more ICL per FLOP than equivalent windowed attention.}
\label{fig:yolo_plots}
\end{figure}

In Section~\ref{sec:balance}, we saw that linear attention has better in-context learning than windowed attention. Now, we investigate whether this is also true for weight-state balanced linear attention, by using power attention. Figure~\ref{fig:yolo_plots} shows the progression through training of the ICL curves of two models with balanced weight-state ratios on context length 4096: window-1024 attention and power attention with $p=2$. Both models are based on the RKWV architecture with the attention layer swapped for their respective attention mechanism (see Appendix~\ref{appendix:experimental_details} for full experimental details). In this setting, we see that the power attention architecture has the steepest ICL curve throughout training. Furthermore, as a result of its better in-context learning ability, power attention outperforms (per FLOP) an equivalent transformer in this setting (see Appendix~\ref{appendix:compute_optimal}).

\subsection{Factors impacting in-context learning}
\label{sec:icl_factors}
\begin{figure}
    \centering
    \begin{subfigure}[t]{0.24\textwidth}
        \includegraphics[width=\linewidth]{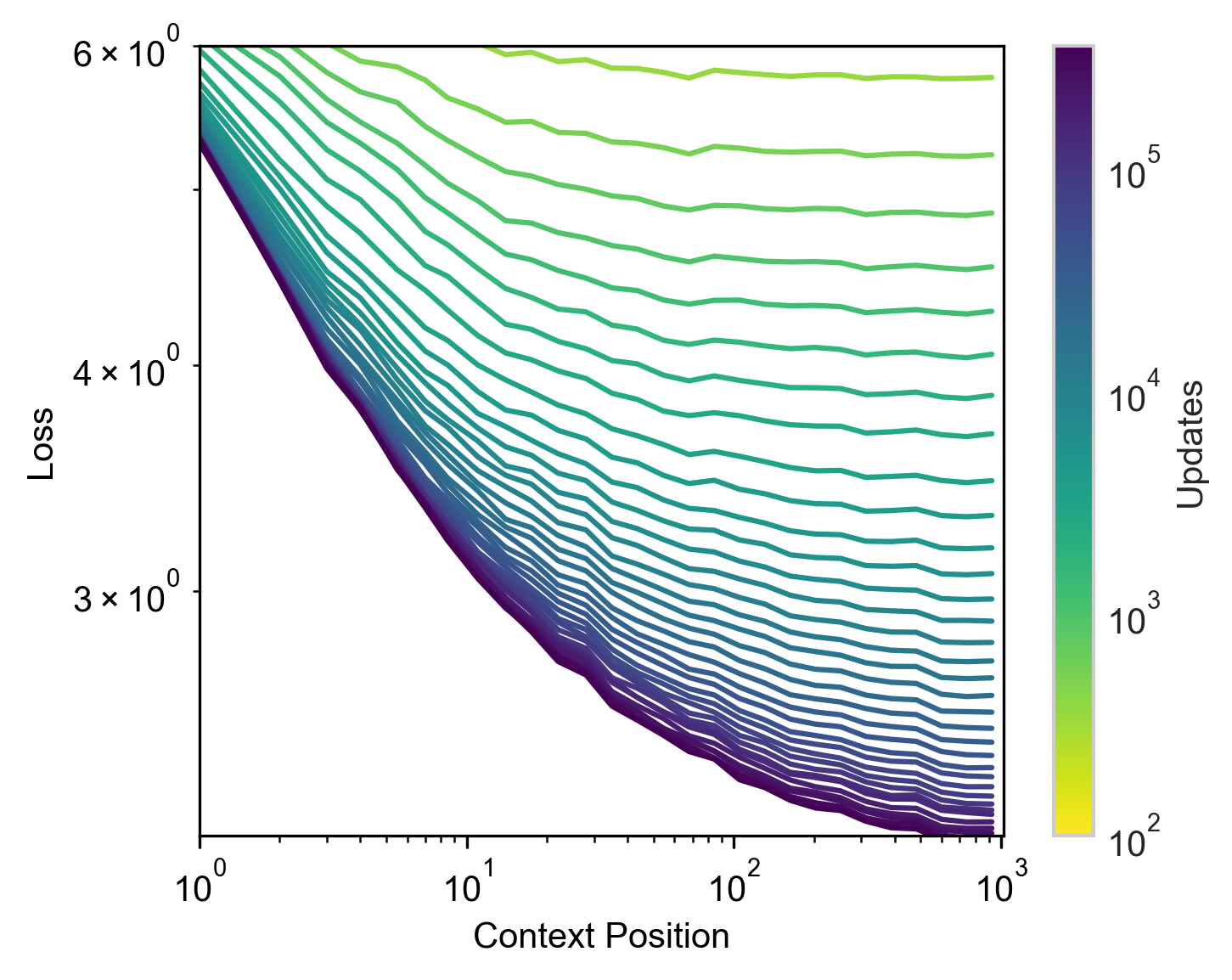}
        \caption{Gradient updates.}
        \label{fig:icl_scaling_iters}
    \end{subfigure}%
    \begin{subfigure}[t]{0.24\textwidth}
        \includegraphics[width=\linewidth]{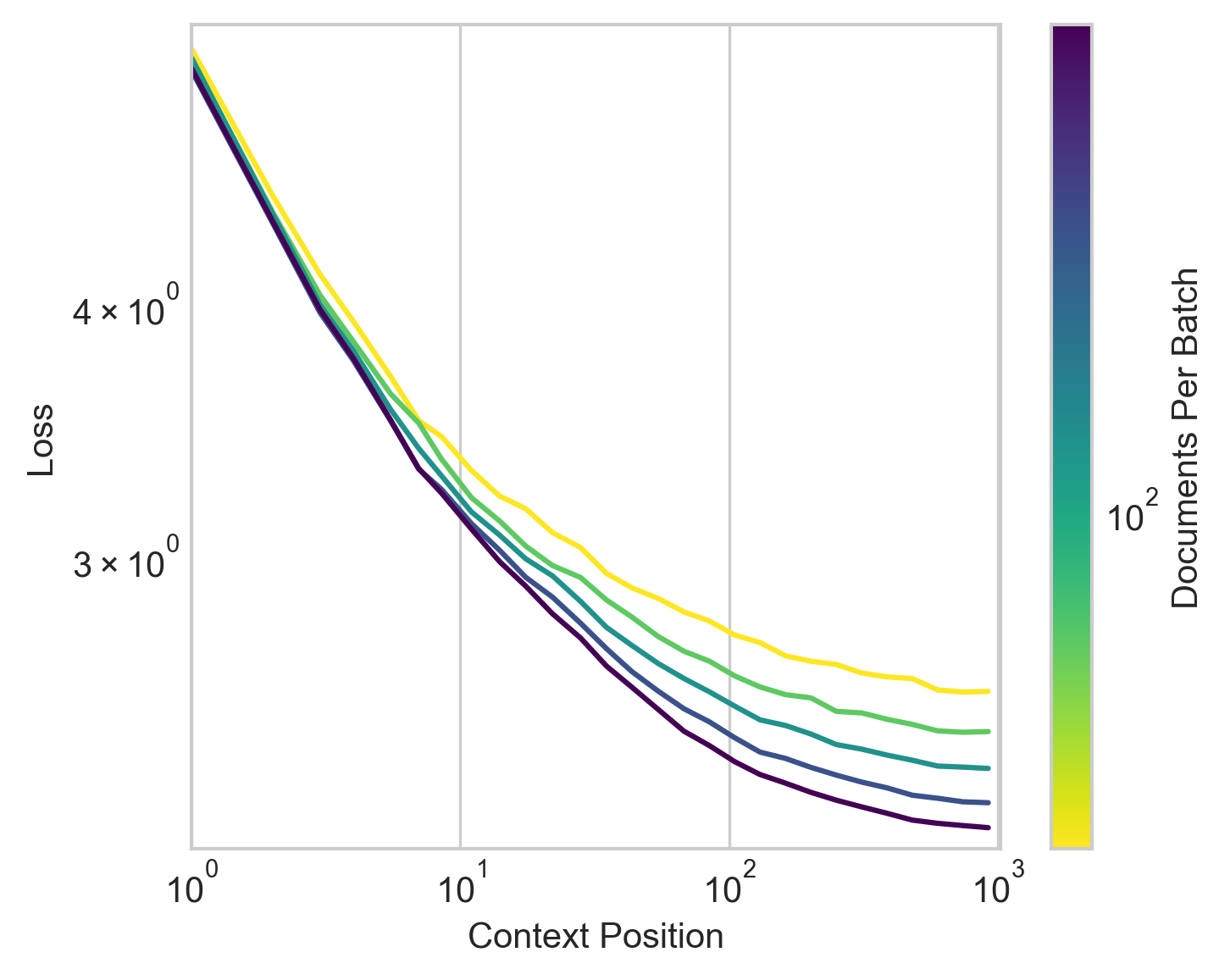}      
        \caption{Documents per batch.}
        \label{fig:icl_scaling_batch}
    \end{subfigure}
    \begin{subfigure}[t]{0.24\textwidth}
        \includegraphics[width=\linewidth]{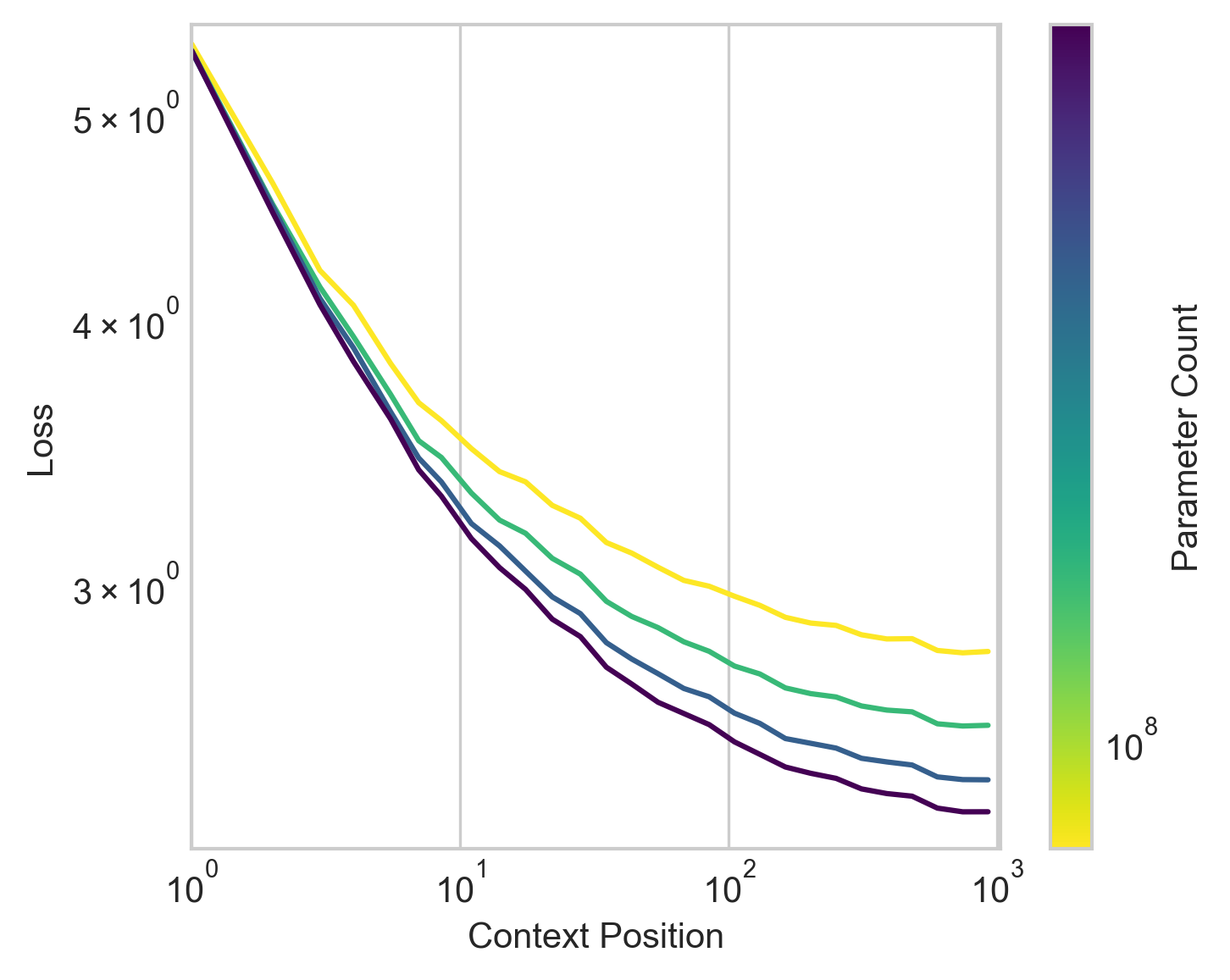}
        \caption{Parameter count.}
        \label{fig:icl_scaling_params}
    \end{subfigure}%
    \begin{subfigure}[t]{0.24\textwidth}
        \includegraphics[width=\linewidth]{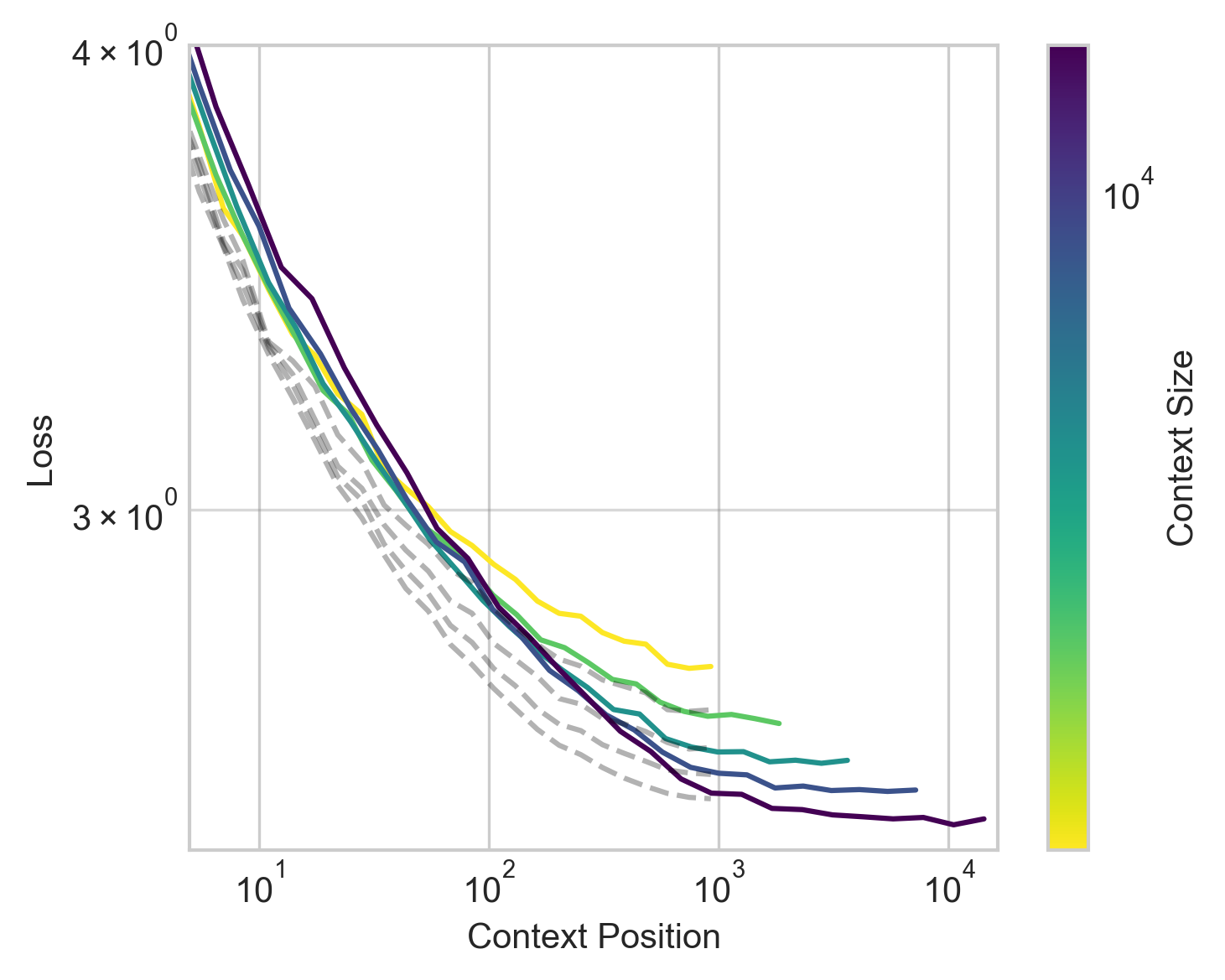}
    \caption{Context length.}
    \label{fig:fooled_by_context_mini}
    \end{subfigure}%
    \caption{The impact of conventional scaling axes on in-context learning of power attention.}
    \label{fig:icl_scaling}
\end{figure}

Figure \ref{fig:icl_scaling} shows the context-wise loss curve of a $p=2$ power transformer when varying four axes: number of gradient updates, documents per batch, parameter count, and context length. See Appendix~\ref{appendix:experimental_details} for experimental details. In all cases, the ICL curve becomes steeper as we scale the respective axis. This indicates that long-context predictions benefit more from scale than short-context predictions.

One phenomenon of note is that scaling context, as shown in Figure~\ref{fig:fooled_by_context_mini}, has two effects: additional opportunity for ICL and additional tokens-per-update (this second effect is similar to that of scaling the batch size). The dashed grey lines on this plot ablate these two factors by sampling long sequences but reshaping into a larger batch of shorter sequences, which removes the effect of ICL and so isolates the effect of the additional tokens. We see that the additional tokens are responsible for nearly all of the improvement, and so the same effect could have been achieved by scaling the batch size.
\footnote{Although we do not explore it in depth in this work, we note that increasing the batch size typically increases the \textit{diversity} of the tokens more quickly than increasing the context length does. This translates into better gradient estimates and improved learning, including improved in-context learning.}
The takeaway is that increasing the context length is not always the best way to improve the in-context learning ability of a model, since \textit{all} axes of scale improve in-context learning.

\subsection{Long-context training}
\label{sec:long_context_training}

\begin{figure}
  \vspace{0pt}
  \centering
  \begin{subfigure}[t]{0.65\textwidth}
    \includegraphics[width=\linewidth]{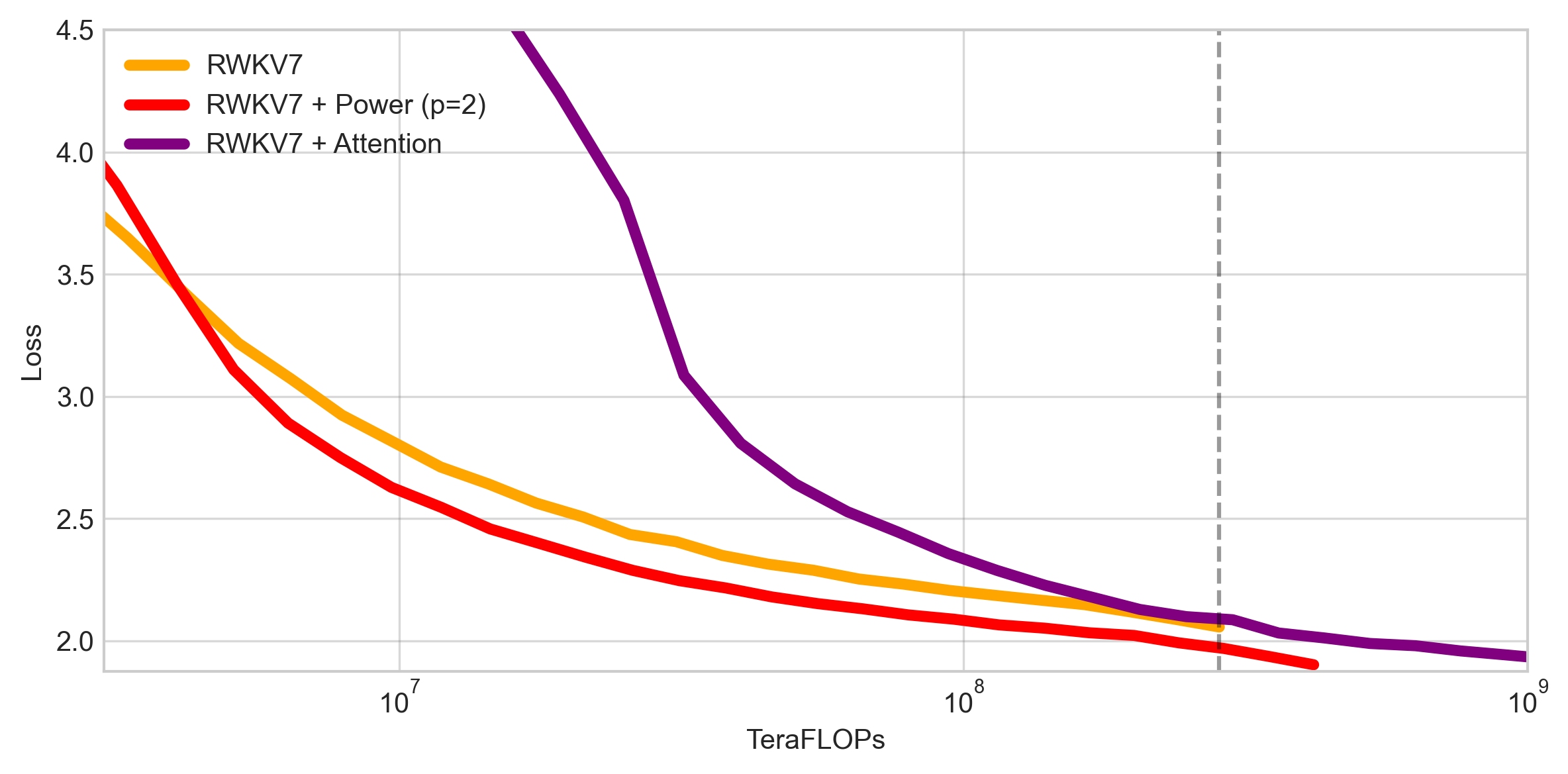}
    \caption{Heldout best-context loss across training.}
    \label{fig:long_context_flops}
  \end{subfigure}\hfill
  \begin{subfigure}[t]{0.32\textwidth}
    \includegraphics[width=\linewidth]{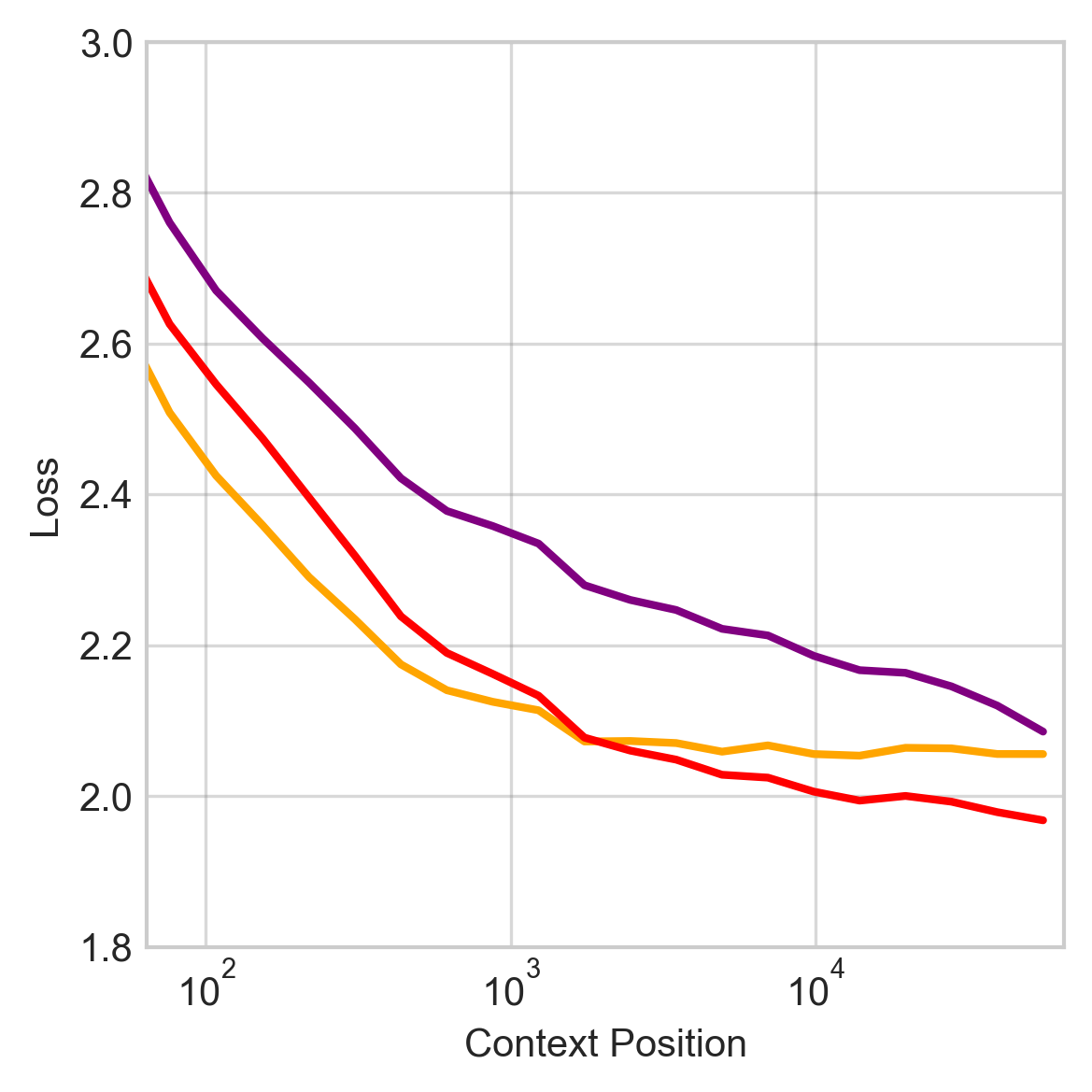}
    \caption{ICL after 3e8 TeraFLOPs.}
    \label{fig:long_context_icl}
  \end{subfigure}
  \caption{Comparison between different forms of attention on long context. The dashed line in \ref{fig:long_context_flops} indicates the position of ICL measurement in \ref{fig:long_context_icl}.}
  \label{fig:long_context}
\end{figure}

We now turn to the question of what architecture is best for training on long contexts. We compare three architectures, all based on RWKV, but with different attention layers: native RWKV linear attention, classic exponential attention, and $p=2$ power attention. Updates are computed on batches of 32 documents, each of length 65536. See Appendix~\ref{appendix:experimental_details} for experimental details. We see in Figure~\ref{fig:long_context_flops} that power attention dominates other architectures in terms of loss-per-FLOP.

The gap in performance between power attention and exponential attention can be attributed to the difference in cost: exponential attention is much more expensive than power attention on long contexts, so the power attention architecture has the opportunity to perform many more steps of gradient descent. In contrast, the gap in performance between power attention and the original RWKV can by attributed to the difference between their in-context learning abilities, as seen in Figure~\ref{fig:long_context_icl}.
RWKV obtains almost no benefit from additional context beyond 2000 tokens.
Power attention allows RWKV to in-context learn nearly as well as exponential attention, while still retaining a large advantage in cost.

We note some limitations of this result. Firstly, a context length of 65536 is far larger than is compute-optimal in this setting, meaning that the dominance of power attention demonstrated here does not directly motivate its use on this dataset. Secondly, note that while power attention dominates in the FLOP regime of our experiments,\footnote{Our $10^9$ TeraFLOPs corresponds to about 1000 H100-hours at 30\% flop utilization.} we expect that given sufficient training FLOPs, the attention model would overtake the power attention model, thanks to its larger state size.


\section{Conclusions \& future work}
\label{sec:conclusion}

Our results indicate that linear attention with a hardware-efficient state expansion is the most effective architecture on long-context training, thanks to state-weight balance and strong in-context learning.
We have proposed power attention, which is one such approach, and hope future architectural research will continue to study a variety of attention variants and state expansions.
For example, one limitation of power attention as currently proposed is its use of the normalization from \citet{attention_is_all_you_need}, which requires positive inner products. This means only even powers are supported, and so the parameter-free adjustments to the state size enabled by adjusting $p$ are coarse.

As discussed in Section~\ref{sec:state_weight_balance} there are many techniques in the literature which reduce the state size of a transformer, including hybrid models, sparse attention, multi-query attention, and latent attention. We investigated one such approach, windowed attention, and found its in-context learning abilities to be worse than linear attention models of the same state size. In the future, a more comprehensive comparison to existing methods would be valuable. Furthermore, a complete characterization of the performance of these algorithms merits rigorous investigation under the framework of scaling laws \citep{scaling_law_for_lm}. Future work should quantitatively explore the impact of state size, context size, and in-context learning on model performance, with the aim of fitting scaling laws dependent on these factors.

Our initial implementation uses Triton \citep{triton}, a high-level tool for writing GPU kernels that allows quick, Pythonic prototyping. However, without the flexibility provided by CUDA, our kernels cannot be optimized as thoroughly as e.g. Flash Attention \citep{flash2}. As a result, our implementation, Power Attention, is not yet as dominant in wall-clock comparisons as FLOPs comparisons would indicate. Future implementations of Power Attention will move from Triton to CUDA in order to push wall-clock performance further.

Our experiments are limited to measuring negative log likelihood on a dataset of generic natural language text. We did not study other domains, modalities, or downstream tasks. In the future, we hope to validate our findings in these settings. Furthermore, we have observed that autoregressive prediction of natural language is largely dominated by short-context dependencies, even on long documents. This diminishes the value of long-context training in this setting. In future work, we hope to discover domains which are dominated by long-term dependencies. For example, we plan to explore tasks that require chain-of-thought reasoning, tool use, and modalities such as audio and video. In domains where performance is heavily dependent on long-term dependencies, the compute-optimal context will be large, and we expect that the dominance of power attention on long contexts will be of practical importance.


\begin{ack}
We would like to thank SF Compute for their generous support on computational resources for this research, Warfa Jibril for data and engineering contributions related to this project, and to Edward Hu, David Brandfonbrener, Eren Malach, and Zhixuan Lin for providing us with feedback on an early draft of this paper.
\end{ack}



\bibliography{main}


\appendix

\section{Derivation of chunked algorithm}
\label{appendix:chunk_derivation}

Here we we prove that the \textit{chunked form} of linear attention is equivalent to the \textit{attention form}. Recall that the chunk-form says
\begin{align}
Y_{(i)_c} = S_{ci}Q_{(i)_c} + V_{(i)_c}\left ( Q_{(i)_c} K_{(i)_c}^T \odot M \right)
\;\;\;\;\;\;\;\;\;\;\;\; 
S_{c(i+1)} = S_{ci} + V_{(i)_c} K_{(i)_c}^T 
\end{align}

Because this is in matrix form, if we look at output at each position $i$, it becomes
\begin{align}
Y_{i} &= S_{ci}Q_i + \sum_{j=\lfloor\frac{i}{c}\rfloor c +1}^i(Q_i K^T_j)V_j \\
&=(S_{c(i-1)} + V_{(i)_c}K_{(i)_c}^T)Q_i + \sum_{j=\lfloor\frac{i}{c}\rfloor c +1}^i(Q_i K^T_j)V_j \\
&=S_{c(i-1)}Q_i + \sum_{j=\lfloor\frac{i-1}{c}\rfloor c +1}^{\lfloor\frac{i}{c}\rfloor c} V_jK_{j}^T Q_i+ \sum_{j=\lfloor\frac{i}{c}\rfloor c +1}^i(Q_i K^T_j)V_j \\
&=S_{c(i-1)}Q_i + \sum_{j=\lfloor\frac{i-1}{c}\rfloor c +1}^{\lfloor\frac{i}{c}\rfloor c} (K_{j}^T Q_i)V_j+ \sum_{j=\lfloor\frac{i}{c}\rfloor c +1}^i(Q_i K^T_j)V_j \\
&=S_{c(i-1)}Q_i + \sum_{j=\lfloor\frac{i-1}{c}\rfloor c +1}^{\lfloor\frac{i}{c}\rfloor c} (Q_iK_{j}^T)V_j+ \sum_{j=\lfloor\frac{i}{c}\rfloor c +1}^i(Q_i K^T_j)V_j \\
&=S_{c(i-1)}Q_i + \sum_{j=\lfloor\frac{i-1}{c}\rfloor c +1}^i(Q_i K^T_j)V_j \\
\vdots \\
&=S_0 Q_i + \sum_{j=\lfloor\frac{0}{c}\rfloor c +1}^i(Q_i K^T_j)V_j \\
&= \sum_{j=1}^i(Q_i K^T_j)V_j \quad\text{assuming initial state is 0}
\end{align}

This concludes the proof.

\section{Derivation of $\spow$}
\label{sec:derivation_of_spow}

\subsection{Tensor product and tensor power} A convenient way to define the tensor product is, given vectors $x,y \in \R^d$, their tensor product $x\otimes y = x y^T \in \R^{d\times d}$. The generic tensor product of $p$ vectors in $\R^d$ can be written as $T = \bigotimes_{k=1}^p x_k \in \R^{d\times \cdots \times d }$ where, evaluated at a multi-index $(i_1 \cdots i_p) \in \mi$, the tensor $T$ has value $T_i = \prod_k x_{k,i_k}$. (For example, if $T= a\otimes b \otimes c$ then $T_{1,2,3} = a_1 b_2 c_3$.)

In this work, a central focus is on the $p$th \textit{tensor power}, defined as taking the tensor product of a vector with itself $p$ times, which we denote using $x^{\otimes p}$ . We can define the helpful $\tpow(x,p)  = \text{flat}\left(x^{\otimes p}\right) \in \R^{d^p}$, which gives us the flattened tensor power as a vector.
\begin{align}
\tpow(x, p) =  
\begin{bmatrix}
x_1 \cdots x_1 \\
 x_1 \cdots x_2 \\
\vdots \\
 x_d \cdots x_d \\
\end{bmatrix}
=
\begin{bmatrix}
\vdots \\
 \prod_k x_{i_k} \\
\vdots \\
\end{bmatrix}_{(i_1, \cdots, i_p)\in \mi}
\end{align}
The central property that makes $\tpow$ useful to us is:
\begin{align}
\tpow(x,p)^T \tpow(y,p)
&= \sum_{(i_1,\cdots) \in \mi} x_{i_1} \cdots x_{i_p}  y_{i_1} \cdots y_{i_p} \\
&= \sum_{i_1 \in \N_d}  x_{i_1} y_{i_1}  \sum_{i_2 \in \N_d}  x_{i_2} y_{i_2}\ \cdots \\
&= (x^T y)^p
\end{align}
Thus, letting $\phi = \tpow$ we see that power attention can be expressed as a special case of linear attention, $Y_i^{\text{attn}_\text{pow}^p} = Y_i^{\text{attn}_\text{lin}^{\tpow(\cdot, p)}}$. Power attention therefore inherits all of the desirable properties of linear attention described in Section 2, including a constant-size state and parallelizable chunked form. $\tpow$ is a state expansion, mapping keys and queries into $\mathbb{R}^{d^p}$, and so power attention possesses a state of size $d^pv$.

\subsection{Symmetric power}

Here we prove that symmetric power $\spow$ is a mathematically equivalent state expansion function to $\tpow$.

Recall from Lemma \ref{lemma:spow_inner_product} that 
\begin{align}
\spow_p(x) = \begin{bmatrix}
    \vdots\\
    \sqrt{\frac{p!}{\hist_k(i)!}}\prod_k x_{i_k} \\
    \vdots
\end{bmatrix}_{i \in NDMI_d^p}
\end{align}

Where each $i = (i_1,...,i_p)$ is the set of non-decreasing-multi-indices that determines a given entry in the embedded vector. One can use a different set of multi-indices $\alpha = (\alpha_1,...,\alpha_d)$ to represent the same embedding, where 
\begin{align}
\alpha_j = \begin{cases}
1 & \text{if }\exists k \in \{1,2,...,p\}, \text{s.t.} i_k = j \\
0 & \text{otherwise}
\end{cases}
\end{align}
in other words, we can include all the dimensions of $x$ in each entry of the expanded vector, and mask out unnecessary dimension by raising them to a power of 0.

With this setup, let $x,y\in \mathbb R^d, \spow_p(x) \in \mathbb{R}^{\binom{d+p-1}{p}}$ be the symmetric power embedding indexed by the multi-indixes $\mathbf{\alpha} = (\alpha_1, ... \alpha_d)$, satisfying $\sum_{i=1}^d \alpha_i = p$.
\begin{align}
\spow_p(x)_{\mathbf \alpha} = \sqrt{\frac{p!}{\alpha_1!\cdots \alpha_d!}}\;x_1^{\,\alpha_1}\cdots x_d^{\,\alpha_d}.
\end{align}
Then, by the multinomial theorem
\begin{align}
\bigl\langle \spow_p(x),\,\spow_p(y)\bigr\rangle
   &=\sum_{\mathbf \alpha}\frac{p!}{\alpha_1!\cdots \alpha_d!}\;
      x_1^{\,\alpha_1}\cdots x_d^{\,\alpha_d}\;
      y_1^{\,\alpha_1}\cdots y_d^{\,\alpha_d} \\
& = \sum_{\mathbf \alpha}\frac{p!}{\alpha_1!\cdots \alpha_d!}
      (x_1 y_1)^{\alpha_1}\cdots(x_d y_d)^{\alpha_d}\\
   &=\bigl(x_1y_1+\dots+x_dy_d\bigr)^{p}\\
   &=(x^T y)^p
\end{align}
Therefore
\begin{align}
    \langle\spow_p(x), \spow_p(y)\rangle = (x^T y)^p = \langle\tpow_p(x), \tpow_p(y)\rangle
\end{align}

\section{LongCrawl64}
\label{appendix:longcrawl}

\begin{figure}[h]
    \centering
    \begin{subfigure}[t]{0.5\textwidth}
        \vspace{0pt}
        \includegraphics[width=\linewidth]{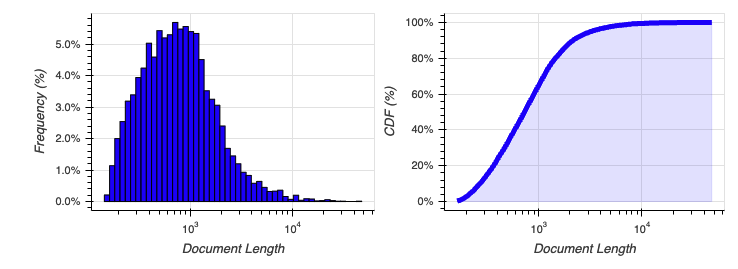}
        \caption{Openwebtext document length distribution.}
    \end{subfigure}%
    \begin{subfigure}[t]{0.33\textwidth}
        \vspace{3pt}
        \includegraphics[width=\linewidth]{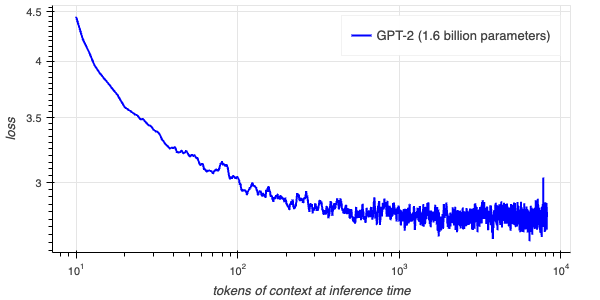}
        \vspace{-8pt}
        \caption{GPT-2 trained with 8192 context.}
    \end{subfigure}
    \caption{}
\end{figure}
\label{fig:long_data}

LongCrawl64 \citep{lc64} consists of 6,661,465 pre-tokenized documents, each of which is 65,536 tokens long, for a total token count of 435 billion. The data is sourced from the Common Crawl, a typical source for language modeling datasets, but pretokenized and filtered down to only include long sequences. This is a prerequisite for in-context learning. In Figure~\ref{fig:long_data}, it is clear that there is little potential for in-context learning beyond the lengths of the majority of documents in the dataset. The figure above and the discussion on the need for long documents were originally presented by \citet{the_one_with_tail_loss}.

\section{Experimental details}
\label{appendix:experimental_details}

Our experiments were implemented in PyTorch \citep{pytorch}, based around the FLA codebase \citep{fla_repo} whose implementations of all architectures we use. Since this work is focused specifically on the attention layer, we typically separate out the architecture (which we consider everything \textit{except} the attention) from the attention itself. Also, since comparing loss between models of different context lengths can be nuanced, we report the best-context loss as described by \citet{the_one_with_tail_loss} whenever plotting a scalar-valued loss for a training curve. When selecting the best context, and also when plotting in-context learning curves, we smooth across the sequence dimension by binning, using exponentially-growing bins so that the bin widths are equal on a log plot.

Unless otherwise noted, we used the following hyperparameters for training: LongCrawl64 (train set for training and report losses on the heldout set), batch size 32, context length 1024, learning rate 3e-4 with a 2000 step warmup from 0 and a cosine decay over the full range of training to 1e-5, one epoch of training (or until convergence), AdamW with weight decay of .1, beta1 of .9, beta2 of .999, gradient clipping of 1, bf16 training, and activation checkpointing for memory reduction. We use a common set of model sizes, taken from \citet{gpt2}. Small models have width 768, 12 hidden layers, 12 heads, and a MLP ratio of 4. Medium models have width 1024, 24 hidden layers, 16 heads, and a MLP ratio of 4. Large models have width 1280, 36 hidden layers, 20 heads, and a MLP ratio of 4. For experiments involving the RWKV architecture, we use RWKV7.

Figure~\ref{fig:lin_vs_exp} and Figure~\ref{fig:icl_comparison_combined} contain a variety of architectures, labeled in the legend; all are small models with default hyperparameters. Figure~\ref{fig:icl_comparison} contains two small RWKV models with default hyperparameters and differing attention layers.

In Figure~\ref{fig:unbalanced_transformers}, all three curves are RWKV + attention models. The model with 1:12 WSFR has 6 heads, 8 layers, head size of 64, hidden dimension of 512, a context of 65536, and a batch size of 32. The model with 4:1 WSFR is a medium model with batch size 512 and context size 4096. The model with 99:1 WSFR has 26 layers, 20 heads, head size of 64, hidden dimension of 1280, batch size 32768, and context size 64. 

Figure~\ref{fig:yolo_plots} and Table~\ref{fig:yolo_table} contain the resuls of experiments run using the RWKV architeture at large (750M) size. The transformer was run using a batch size of 7680 and context of 1024. The other runs used batch size 1920 and context size 4096.

Figure~\ref{fig:icl_scaling_iters} consists of one large $p=2$ transformer. Figure~\ref{fig:icl_scaling_batch} consists of small $p=2$ transformer models evaluated at iteration 64k. Figure~\ref{fig:icl_scaling_params} consists of large $p=2$ transformers evaluated at iteration 64k. The curve corresponding to the smallest model scale has 512 width, 6 layers, and 8 heads; the other three are small, medium, and large models. 
Figure~\ref{fig:fooled_by_context_mini} involves a sweep over context lengths on a $p=2$ transformer with 512 width, 6 layers, and 8 heads, evaluated at 170k iterations. The grey curves are the same, but after sampling documents (of any length), the data is reshaped to have a block size of 1024 before training. This means the actual tokens are kept constant, but the model has no opportunity to use long context to learn.

Figure~\ref{fig:long_context_flops} shows three small RWKV models with different attention layers, trained on context length of 65536 tokens at a batch size of 32.

All experiments in this work were run on Nvidia H100 GPUs, typically on nodes of 8 GPUs. In the course of conducting the experimental portion of this work, we had access to between 32 and 300 H100s for two months. The majority of the compute was spent on research, with only about 20\% of compute spent on experiments in this paper.

\section{Power attention is compute-optimal under inference latency constraints}
\label{appendix:compute_optimal}

To a first-order approximation, inference latency is proportional to the sum of the parameter count and the state size. This is relevant when choosing an optimal training strategy for a model whose ultimate usage will have inference constraints, for example a speech model model \citep{wang2021deep}. A reasonable approach is to choose the parameter count and state size to be at some tolerable scale, and spend the compute budget scaling other axes of training, such as batch size and context length.

\begin{wraptable}{r}{0.55\textwidth}

\centering
\begin{tabular}{lcc}
\toprule
 & Best context length & Loss \\
\midrule
Window (1k) & 1878 & 1.638 \\
Attention & 1024 & 1.631 \\
Power ($p=2$) & 4096 & \textbf{1.613} \\
\bottomrule
\end{tabular}
\caption{Power attention is compute-optimal given sufficient train FLOPs, when inference latency is equal.}
\label{fig:yolo_table}

\end{wraptable}

In this setting, we can compare the best achievable performance of a transformer to that of other models, keeping all variants equivalent in terms of state size, parameter count, tokens per update, and total FLOPs. In Table \ref{fig:yolo_table}, we construct three such models, and compare the final best-context loss. The transformer is trained on batches of 7680 document of length 1024 (a context length which keeps its state size equivalent to the other approaches), while the windowed transformer and power attention transformer use batches of 1920 documents of length 4096. Power attention is the only architecture to outperform the transformer. See Appendix~\ref{appendix:experimental_details} for experimental details.

\section{Algorithms in power attention}
\label{sec:algorithms}

In this section we present the four algorithms used in power attention.

\subsection{Attention}
As shown below, the \textit{attention} kernel in Power Attention is very similar to Flash Attention, apart from an extra step of log-space power and an extra output for the normalization term $l$. We chose to raise the attention score matrix to a power $p$ in the log space because it is more numeric stable than performing the power operation directly.

\begin{table}[htbp]
\centering
\begin{tabular}{@{}l l l@{}}
\toprule
\textbf{Step} & \textbf{Flash Attention} & \textbf{Power Attention (attention form)}\\
\midrule
1. Query-Key Inner Product  & $S = QK^T$ & $S = QK^T$\\
2. Softmax Scaling & $S = S\odot \text{scale}$ & $S = S\odot \text{scale}$ \\
3. Log-space Power & & $S = p \log(|S| + \epsilon) $ \\
4. Row max scaling & $S = S - \text{rowmax}(S)$ & $S = S - \text{rowmax}(S)$ \\
5. Masked Exponential &$P = \text{exp}(S \odot M)$& $P = \text{exp}(S\odot M)$\\
6. Normalization &$P = P \odot D^{-1}(\text{rowsum}(P))$& $\zeta = \text{rowsum}(P)$ \\
7. Matmul with Value & $O = PV$ & $O = PV$ \\
8. Output & $O$ & $O, \zeta$\\
\bottomrule
\end{tabular}
\caption{Procedural comparison between Flash Attention and Power Attention (attention form). $Q, K \in R^{t\times d}$, $V \in R^{t\times v}$; $p$ stands for the degree of power; $\epsilon$ is a small constant to avoid taking the log of zero; rowmax(P) refers to the operation of taking the max of the $t\times t$ attention score matrix, an often-used techniques for stabilizing softmax \citep{online_softmax}; $\text{rowsum}(P)$ refers to the operation of summing up each row of the softmax matrix; $D^{-1}$ refers to the operation of converting a vector into a diagonal matrix and take its inverse; $\zeta$ is the normalization term (sum of attention scores) used for combining attention output and \textit{query-state} output}
\end{table}

\subsection{Update state}
The \textit{update-state} operation concerns with creating a new state $S_{i+1} \in R^{D\times v}$ based on the past state $S_{i} \in R^{D\times v}$ and all the keys $K_{i} \in R^{c\times d}$ and values $V_{i} \in R^{c\times v}$ in the current chunk. There are many variants to this formulation in modern RNNs. Specifically, past state $S_i$ are usually gated with a decay factor $\gamma_i$, which often depend on input as well. 
\begin{align}
\text{update-state}(S_{i}, K_{i}, V_{i}) &= S_i + \phi(K_i)^TV_i \\
\text{gated-update-state}(S_{i}, K_{i}, V_{i}) &= S_i\odot \gamma_i + \phi(K_i)^TV_i
\end{align}
Regardless the exact state evolution formula, the fusion of state expansion and a subsequent matrix multiplication is the fundamental building block, which we termed \textbf{fused spow-mma (expand M)} kernel. We use \textbf{expand M} here as \textbf{M, N, K} are commonly used to denote the 3 dimensions of a matrix multiplication problem, and this kernel is expanding the state along \textbf{M} axis ($K_i^T \in R^{d\times c} \rightarrow \phi(K_i)^T\in R^{D\times c}$). We might also use the term \textit{update-state} interchangeably with \textbf{fused spow-mma (expand M)} kernel, as the gated summation is done in the \textit{discumsum} kernel. Note that in practice, the \textit{update-state} kernel would also produce a normalization term (a.k.a. sum of expanded keys) $\gamma$, which is used to combine the outputs of chunked attention and \textit{query-state} such that the output is normalized. We denote this kernel with \textit{fused-update-state}.
\begin{align}
\text{fused-update-state}(S_i, K_i, V_i) &= (\phi(K_i)^TV_i, \phi(K_i)^T\mathbf{1})
\end{align}

\begin{algorithm}[H]
\caption{Fused Update State}
\begin{algorithmic}[1]
\Require Matrices $A$ of size $\mathbf{d} \times \mathbf{K}$, $B$ of size $\mathbf{K} \times \mathbf{N}$
\Ensure Output matrix $C$ of size $\mathbf{D} \times \mathbf{N}$, normalization factor $\gamma$ of size $\mathbf{D}$
\State Define degree of power $\mathbf{p}$, tile size for expansion $\mathbf{d_{tile}}$, expanded tile size $\mathbf{D_{tile}} = \mathbf{d_{tile}}^{\mathbf{p}}$
\State Denote the ordered list of non-decreasing-multi-indices $\mathbf{NDMI}^{\mathbf{p}}_{\mathbf{d}/\mathbf{d_{tile}}}$ with $\mathbf{\lambda}$, of size $\mathbf{L} \times \mathbf{p}$
\State Divide A into $N_A = \lceil\mathbf{\frac{K}{TK}}\rceil$ tiles, $A_1, ..., A_{N_A}$, each of size $\mathbf{d} \times \mathbf{TK}$; divide each $A_k$ further into $\mathbf{N_d = \frac{d}{d_{tile}}}$ subtiles, $A_k^1, ..., A_k^{\mathbf{d/d_{tile}}}$, each of size $\mathbf{d_{tile} \times TK}$
\State Divide B into $N_B = \lceil\mathbf{\frac{N}{TN}}\rceil$ tiles, $B_1, ..., B_{N_B}$, each of size $\mathbf{K} \times \mathbf{TN}$; divide each $B_j$ further into $\lceil\mathbf{\frac{K}{TK}}\rceil$ subtitles, $B_j^1, ..., B_j^{\lceil \mathbf{K/TK}\rceil}$, each of size $\mathbf{TK} \times \mathbf{TN}$
\For{ $1 \leq l \leq \mathbf{L}$, in parallel }
    \For{ $1 \leq j \leq N_B$, in parallel }
        \State Initialize accumulation registers: $C_{l,j} \gets 0$ of shape $\mathbf{D_{tile} \times TN}$, D
        \State Initialize register for matrix multiplication $\hat{A}_k \gets 0$ of shape $\mathbf{D_{tile}}\times \mathbf{TK}$
        \State Initialize register for normalization factor: $\gamma_l$ of shape $\mathbf{D_{tile}}$
        \For{$1 \leq k \leq \lceil\mathbf{\frac{K}{TK}}\rceil$}
          \State Load $A_{k}$ from global memory to on-chip SRAM
          \State Load $B_{j}^k$ from global memory to on-chip SRAM
          \For{$1 \leq z \leq \mathbf{p} $}
          \State Load $A_{k}^{\mathbf{\lambda}(l, z)}$ from on-chip SRAM into registers
          \EndFor
          \State $\hat{A_k} \gets A_k^{\mathbf{\lambda}(l,1)} \otimes\cdots\otimes A_k^{\mathbf{\lambda}(l, \mathbf{p})}$
          \State $\gamma_l \gets \text{rowsum}(\hat{A^k}) + \gamma_l$
          \State $C_{l,j} \gets \hat{A_k} B_j^k + C_{l,j}$
        \EndFor
        \State Write $C_{l,j}, \gamma_l$ to global memory
    \EndFor
\EndFor
\end{algorithmic}
\end{algorithm}

\subsection{Discumsum}
The \textit{discumsum} operation involves discounting and accumulative-summing states $S \in R^{n\times D\times d}$ for each chunk in a sequence (hence the name), where $n = \lceil\frac{t}{c}\rceil$. It takes the output produced by \textit{update-state} kernel, and a gating factor $\lambda \in R^n$ and produced the discounted accumulative sum. Discounting is necessary when gating is involved. The discumsum kernel used in in paper was implemented by a custom CUDA kernel.

\begin{align}
\text{discumsum}(S,  \lambda) &= 
\begin{bmatrix}
S_1\\
S_1 \odot \lambda_1 + S_2 \\
\cdots \\
S_1 \odot \lambda_1  + \cdots + S_i\odot\prod_{j=1}^i\lambda_j + S_n
\end{bmatrix}
\end{align}

\subsection{Query state}
The \textit{query-state} kernel involves querying the past state $S_i$ using the queries $Q_i\in R^{c\times d}$ in the current chunk.
\begin{align}
\text{query-state}(S_i, Q_i) = \phi(Q)S_i
\end{align}

Notice that as opposed to the \textit{update-state} kernel, the \textit{query-state} kernel expands the queries along the dimension of reduction in matrix multiplication. Therefore in the inner loop of the kernel, we go through all the nondecreasing-multi-indices. 

In practice, we also fuse the summation of the intra-chunk output from attention $Y\in R^{c\times v}$ and $\phi(Q)S_i$ into the \textit{query-state} kernel itself. We also chose to fuse the normalization into it. The algorithm for \textit{fused-query-state} is shown below.
\begin{align}
\text{fused-query-state}(S_i, Q_i, Y_i, \zeta_i, \gamma_i) = \frac{Y_i + \phi(Q_i)S_i}{\zeta_i + \phi(Q_i)\gamma_i}
\end{align}

\begin{algorithm}[H]
\caption{Fused Query State}
\begin{algorithmic}[1]
\Require Matrices $A$ of size $\mathbf{M} \times \mathbf{d}$, $B$ of size $\mathbf{D} \times \mathbf{N}$, $Y$ of size $\mathbf{M} \times \mathbf{N}$, $\gamma$ of size $\mathbf{D}$, $\zeta$ of size $\mathbf{M}$
\Ensure Output matrix $C$ of size $\mathbf{M} \times \mathbf{N}$
\State Define degree of power $\mathbf{p}$, tile size for expansion $\mathbf{d_{tile}}$, expanded tile size $\mathbf{D_{tile}} = \mathbf{d_{tile}}^{\mathbf{p}}$
\State Denote the ordered list of non-decreasing-multi-indices $\mathbf{NDMI}^{\mathbf{p}}_{\mathbf{d}/\mathbf{d_{tile}}}$ with $\mathbf{\lambda}$, of size $\mathbf{L} \times \mathbf{p}$
\State Divide A into $N_A = \lceil\frac{M}{TM}\rceil$ tiles, $A_1, ..., A_{N_A}$, each of size $\mathbf{TM} \times \mathbf{d}$; divide each $A_i$ further into $N_d = \frac{\mathbf{d}}{\mathbf{d_{tile}}}$ subtitles, $A_i^1, ..., A_i^{N_d}$, each of size $\mathbf{TM}\times\mathbf{d_{tile}}$
\State Divide B into $N_B = \lceil\frac{N}{TN}\rceil$ tiles, $B_1, ..., B_{N_B}$, each of size $\mathbf{D} \times \mathbf{TN}$; divide each $B_i$ further into $\mathbf{L}$ subtitles, $B_i^1, ..., B_i^{\mathbf{L}}$, each of size $\mathbf{D_{tile}} \times \mathbf{TN}$
\State Divide $Y$ into $N_A$ tiles, $Y_1,...Y_{N_A}$, each of size $\mathbf{TM} \times \mathbf{N}$; divide each $Y_i$ further into $N_B$ subtiles, $Y_i^1,...,Y_i^{N_B}$, each of size $\mathbf{TM}\times \mathbf{TN}$
\State Divide $\gamma$ into $\mathbf{L}$ tiles, $\gamma_1,...\gamma_L$, each of size $\mathbf{D_{tile}}$; divide $\zeta$ into $N_A$ tiles, $\zeta_1, ...,\zeta_{N_A}$, each of size $\mathbf{TM}$
\For{ $1 \leq i \leq N_A$, in parallel } 
    \For{ $1 \leq j \leq N_B$, in parallel }
        \State Initialize accumulation registers: $C_{i,j} \gets 0$ of shape $\mathbf{TM \times TN}$
        \State Initialize register for matrix multiplication $\hat{A}_i \gets 0$, of shape $\mathbf{TM}\times \mathbf{D_{tile}}$
        \State Initialize register for normalization $s \gets 0$, of shape $\mathbf{TM}$
        \State Load $A_{i}$ from global memory to on-chip SRAM
        \For{$1 \leq l \leq \mathbf{L}$}
          \State Load $B_{j}^l, \gamma_l$ from global memory to on-chip SRAM
          \For{$1 \leq z \leq \mathbf{p} $}
          \State Load $A_{i}^{\mathbf{\lambda}(l, z)}$ from on-chip SRAM into registers
          \EndFor
          \State $\hat{A_i} \gets A_i^{\mathbf{\lambda}(l,1)} \otimes\cdots\otimes A_i^{\mathbf{\lambda}(l, \mathbf{p})}$
          \State $s \gets \hat{A_i} \lambda_l + s$
          \State $C_{i,j} \gets \hat{A_i} B_j^l + C_{i,j}$
        \EndFor
        \State Load $Y_i^j, \zeta_i$ from global memory to on-chip SRAM
        \State $C_{i,j} \gets \frac{Y_i^j + C_{i,j}}{\zeta_i + s}$
        \State Write $C_{i,j}$ to global memory
    \EndFor
\EndFor
\end{algorithmic}
\end{algorithm}

\newpage

\end{document}